%% file: main.tex
\documentclass[acmsmall,nonacm]{acmart}
\setcopyright{none}
\input{packages}
\begin{document}

\title{Intent Representation and Human Distraction Modulate the Effects of Legible Robot Motion in Hallway Navigation}
\author{Pranav Goyal}
\orcid{0009-0004-0996-644X}
\affiliation{%
  \institution{University of Michigan at Ann Arbor}
  \city{Ann Arbor}
  \country{USA}
}
\email{prgoyal@umich.edu}

\author{Andrew Stratton}
\orcid{0009-0003-8126-6299}
\affiliation{%
  \institution{University of Michigan at Ann Arbor}
  \city{Ann Arbor}
  \country{USA}
}
\email{arstr@umich.edu}
\author{Christoforos Mavrogiannis}
\orcid{0000-0003-4476-1920}
\affiliation{%
  \institution{University of Michigan at Ann Arbor}
  \city{Ann Arbor}
  \country{USA}
}
\email{cmavro@umich.edu}

\input{abstract}


\maketitle

\input{sections/intro.tex}
\input{sections/relatedwork}
\input{sections/technical.tex}
\input{sections/experiments.tex}
\input{sections/discussion.tex}
\bibliographystyle{ACM-Reference-Format}
\bibliography{references}

\end{document}

%% file: packages.tex
\usepackage{subcaption}
\usepackage{tabularx}

%% file: abstract.tex
\begin{abstract}

Legibility in human-robot interaction (HRI) is often described as the property of robot motion that enables an observer to confidently and quickly infer the robot’s intent. Mature frameworks exist for generating legible motion in front of static observers, typically by making the robot’s destination apparent as early as possible. Social robot navigation in public spaces, however, presents a different challenge: pedestrians are themselves in motion, often with divided attention, and must coordinate with the robot in real time. In such encounters, where the robot is ultimately headed may matter less to a pedestrian than how it intends to pass them. This raises two questions: which aspect of intent should legible motion convey, and does legibility remain beneficial when pedestrians are not fully attentive to the robot? We address these questions in the ubiquitous and demanding scenario of hallway navigation, through two controlled user studies comparing multiple legibility formulations implemented within a shared model predictive control framework. Study~1 ($N=45$) compares destination-based and passing-side intent representations, showing that the latter, particularly when adaptively updated, allows for smoother human motion and is perceived as more competent and less mentally and physically demanding than the former. Study~2 ($N=45$) examines the effect of pedestrian attention, demonstrating that legible motion supports smooth human motion even under distraction, although this benefit is not consistently reflected in subjective ratings. Together, these findings suggest that effective legible motion in social robot navigation benefits from interaction-level intent representations that support coordination, with some effects persisting even when human attention is divided. Code is available at \url{https://github.com/fluentrobotics/Legible_MPPI} and footage from our experiments at \url{https://www.youtube.com/watch?v=P5O5BwfcUCo}.

\end{abstract}

%% file: sections/intro.tex
\section{Introduction}

\begin{figure}[t]
    \centering
    \includegraphics[width = \linewidth]{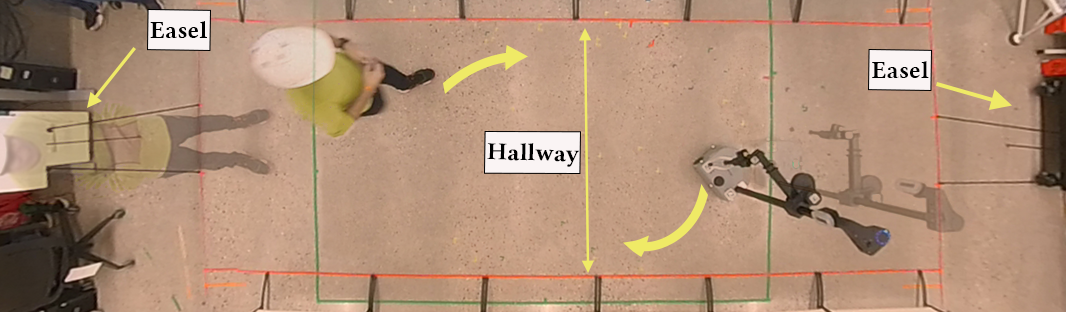}
    \Description{Overhead photograph of the experimental hallway, a rectangular floor area bounded by red tape lines with a labeled easel at each end. A participant in a white helmet and the Stretch~2 robot approach each other from opposite ends, with faded copies showing their earlier positions, and curved yellow arrows show them veering toward opposite walls to pass.}
    \caption{An instance from our study of legible motion for social robot navigation in a hallway. The participant performs a mock inspection task, repeatedly traversing the hallway to place colored stickers on the easel pad at each end, while the Hello Robot Stretch~2 moves in the opposite direction. Each traversal produces a head-on encounter (yellow arrows) in which the participant and robot must coordinate to pass within the constrained space; across trials the robot executes different legible-motion strategies.
    }
    \label{fig:experiment_setup}
\end{figure}

Robots moving in pedestrian spaces like the workplace, the home, and public spaces must do more than simply avoid collisions: they must move in ways that people can intuitively interpret. In social robot navigation (SRN), robots are required to share space with pedestrians and continuously coordinate their motion during close-proximity encounters. In human–human encounters, pedestrians rely on subtle cues such as gaze, orientation, and small path adjustments to infer each other’s intentions and coordinate smooth passages~\citep{Wolfinger95,karp1977being,goffman,goffman2009relations}. Robots, however, lack these ingrained social signals. When their motion is difficult to interpret, pedestrians may hesitate, yield unnecessarily, or experience discomfort, leading to inefficient and sometimes unsafe interactions ~\citep{core-challenges2021}. This motivates the incorporation of intent signaling directly into robot motion control, particularly in settings where humans and robots must continuously coordinate their actions.

One notable way for encoding communicative signals into robot motion in HRI is through the \emph{Legibility} framework of~\citet{Dragan} who define \emph{Legibility} as the property of motion that enables an observer to quickly and confidently infer the robot’s intent. Early work established that legibility can improve collaboration in manipulation tasks by reducing ambiguity and building trust. For example, \citet{Dragan2013GeneratingLM} showed that legible trajectory planning can make a robot’s intended goal easier to infer. This work inspired a large body of work on understanding how legibility shapes HRI under various settings and domains~\citep{Dragan,Dragan2013GeneratingLM,Dragan2015study, kirsch, Busch_experiment,bodden,mavrogiannis2018social,taylor2022observer,FARIA2024104107}. For example, subsequent studies demonstrated that intent-expressive motion enhances perceived safety and efficiency in shared tasks~\citep{Dragan2015study, kirsch, Busch_experiment}. Later work recognized the value of legibility for any joint HRI activity~\citep{knepper_hri_2017}, discussed the impact of important parameters like the observer's viewpoint~\citep{stefanos_visiblity,taylor2022observer}, and proposed frameworks for transferring the benefits of legibility in other domains~\citep{mavrogiannis2018social,kirsch,walker2021corl}.

Despite this rich activity, important limitations arise when legibility is applied to navigation scenarios involving embodied interaction. Much of the existing legibility literature is formulated around simplified settings, in which intent is communicated to passive, static observers who do not spatially interact with the robot~\citep{bodden,Busch_experiment,Dragan}. In such settings (e.g., manipulation tasks), legibility is often framed as enabling an observer to infer a robot’s intended goal or final destination from its motion. However, these formulations often break down in embodied navigation: pedestrians do not need to know where others are ultimately headed; they just need to know how to resolve a possible conflict~\citep{mavrogiannis2018social, Wolfinger95}. Additionally, prior work often evaluates legibility via studies involving passive watching of robot motion~\citep{kim_online_study, szafair_online_study, taylor2022observer,Dragan}, overlooking that in real-world navigation, people are both observers and actors who dynamically adapt to the robot's movements.
Existing work on legibility in navigation has also predominantly emphasized open spaces~\citep{mavrogiannis2022social, taylor2022observer, kirsch}, while constrained environments such as hallways, which force closer proximity and sharper coordination demands, remain comparatively underexplored. Finally, prior work on legible motion assumes that humans are fully attentive observers~\citep{Dragan,Dragan2013GeneratingLM,Dragan2015study, kirsch, Busch_experiment,bodden,mavrogiannis2018social,taylor2022observer,FARIA2024104107}, yet in many real-world environments, including SRN domains, human attention is often divided across multiple tasks like texting, speaking on the phone, carrying objects, etc. These distractions may limit the extent to which robot motion can influence human behavior.

Motivated by the aforementioned gaps in realizing legible SRN, we ask two questions that to the best of our knowledge have not been sufficiently explored. First, we want to understand \emph{how the representation of intent shapes the effectiveness of legible motion in constrained navigation settings}. As mentioned, while prior work emphasizes the communication of the robot's intended destination or an intermediate subgoal, our insight is that navigation under dynamic, crowded, and constrained conditions makes the notion of destination challenging to properly define. Second, we want to understand \emph{how human attention influences the impact of legible motion during interaction}. Prior work emphasizes fully attentive observers, but SRN domains like train stations, airports, and other public places feature a rich context that divides pedestrian attention.

To approach these central questions, we embed alternative formulations of legible motion into a model predictive control framework and evaluate them across two comparative studies in a hallway setting.
Study 1 ($N=45$) explores the impact of intent representation on robot navigation as well as human navigation and impressions. Study 2 ($N=45$) examines the value of legible motion under distraction. Unlike prior work that considers distraction primarily as a robustness check for visibility or timing effects \citep{adathesis}, we treat human attention as a central experimental variable and examine how it shapes the influence of legible motion on behavior and perception. Across both studies, we analyze subjective perceptions of competence, comfort, and effort, alongside objective measures of trajectory smoothness and human acceleration. Together, these studies provide initial empirical insights into key factors influencing the effectiveness of legible motion in SRN, with particular emphasis on intent representation and human attention in dynamic, constrained interactions. In summary, this work makes the following contributions:
\begin{itemize}
    \item We formulate and compare alternative intent representations for legible motion in social robot navigation, contrasting destination-based formulations with interaction-level coordination in constrained environments.
    \item We embed different legibility formulations within a shared model predictive control framework, enabling controlled comparison that isolates the effects of intent representation and adaptability.
    \item We examine human attention as a moderating factor by introducing distraction as an experimental variable and characterizing its impact on the effectiveness of legible motion.
    \item We empirically evaluate legible motion under embodied, close-proximity interaction through two controlled hallway studies, analyzing both human navigation behavior and subjective perception.
\end{itemize}

A short note discussing this work appeared as a four-page Late-Breaking Report (LBR) in the Companion Proceedings of the 21st ACM/IEEE International Conference on Human–Robot Interaction (HRI 2026)~\citep{Goyal2026LegibilityHRICompanion}. The present manuscript discusses the same two user studies but in a more in-depth fashion, motivating the investigation based on an extended literature review, providing a rigorous mathematical formulation, a comprehensive analysis of objective and subjective measures collected during the studies, and extended discussions of our takeaways. In particular, we add a \emph{Foundations} section that formalizes legible social navigation, presenting a unified model predictive control formulation, a model of human intent inference and attention, and an explicit distinction between legibility and predictability. Second, beyond the subjective ratings reported previously, we introduce objective, trajectory-based measures of legibility, namely conflict-resolution time, which captures how early an incoming user commits to a passing side, and robot responsibility, which captures the robot's share of the avoidance, and analyze both across all conditions in both studies. Third, we add a \emph{Discussion} section that develops the design implications of our findings and examines the effect of potential confounds, such as adherence to passing-side conventions, on the observed differences. The journal version further provides fuller statistical reporting, open-ended participant responses, and additional figures illustrating qualitative behavioral differences across conditions.

%% file: sections/relatedwork.tex
\section{Related Work}

We review recent work on SRN and legible motion in HRI, as well as more directly relevant work on legible SRN and our evaluation domain of navigation in hallways.

\textbf{Social robot navigation}.
Earlier work on social robot navigation~\citep{TEB,ORCA} largely treated humans as moving obstacles, missing the inherently interactive nature of human navigation. Interaction-aware methods instead model mutual adaptation, treating humans as intelligent agents perceiving and reacting to the motion of others~\citep{trautmanijrr,mavrogiannis2022social}. This view has been realized both in learning-based policies trained in simulated crowds or from human demonstrations~\citep{sacadrl,cuan2022gesture2pathimitationlearninggestureaware,kathuria2025learningimplicitsocialnavigation} and in model-based planners that fold predicted human motion into the robot's optimization~\citep{SunM-RSS-21,mavrogiannis2023winding}. Beyond avoiding collisions, these methods increasingly seek motion that is efficient, comfortable, and consistent with social norms such as personal space and passing-side conventions~\citep{Wolfinger95,core-challenges2021}. Across these methods, however, interaction is most often framed in a single direction, with the robot inferring human intent in order to resolve conflicts safely and efficiently.

A growing body of work recognizes that navigation is also bidirectional communication, in which the robot's own motion conveys its intent to those around it~\citep{mavrogiannis2022social,core-challenges2021}. When robot motion is easy to interpret, pedestrians coordinate more smoothly and perceive the robot as safer~\citep{core-challenges2021}, motivating methods that not only read human intent but actively express the robot's own~\citep{mavrogiannis2022social,Che_2020}.

\textbf{Legible motion}. Prior work has highlighted the potential of robot intent expressiveness for enabling seamless HRI ~\citep{mallstudy,dugas2020ian,Dragan,mavrogiannis2022social}, through the use of modalities like gestures~\citep{hart2014gesture,dugas2020ian}, lights~\citep{angelopoulos2022familiar}, and motion~\citep{Dragan,Bastarache,taylor2022observer,mavrogiannis2022social,walker2021corl}. The latter, commonly referred to as legible motion, has gained significant attention due to its potential to convey robot intent directly encoded within the robot's otherwise task-oriented motion, without necessarily requiring dedicated signaling components. Formulations of legibility typically frame it as an inference problem: trajectories are planned so that partial observations maximize the observer’s posterior over the robot’s intended goal or action~\citep{Dragan,Dragan2013GeneratingLM,Legibilitydiffuser}. Recent work has further extended this idea to sequential decision-making tasks from single goals ~\citep{FARIA2024104107}. Empirical studies have demonstrated that legible motion improves collaboration by reducing ambiguity, enhancing efficiency, and fostering trust and safety in joint tasks~\citep{Busch_experiment,Dragan2015study,legiblesafety}. However, evaluating legible motion is a challenge in itself. In many studies, users watch non-interactive robot demonstrations, either in person or on video, and infer robot goals, sometimes rating confidence~\citep{Dragan, kim_online_study,szafair_online_study,taylor2022observer,wallkotter2022newapproachevaluatinglegibility}. For example, \citet{adathesis} examined how visibility and timing shape the effectiveness of legible motion, primarily in simulation and static-observer settings. While informative for studying perceptual inference, these paradigms abstract away the reciprocal nature of interaction present in many HRI scenarios, including social robot navigation, where pedestrians both observe and adapt their motion~\citep{trautmanijrr,core-challenges2021}. This highlights the need for embodied experiments in complex scenarios such as narrow hallways, where misunderstandings can escalate into stand-offs~\citep{mavrogiannis2023winding,stratton2024characterizingcomplexitysocialrobot,Fujioka2024}, in contrast to prior work emphasizing relatively open spaces where complexity plays little role~\citep{mavrogiannis2022social,taylor2022observer,kirsch}.

\textbf{Legible motion in social navigation}. Extending legibility~\citep{Dragan2013GeneratingLM,bodden,Busch_experiment} to pedestrian-rich environments introduces unique challenges, as social navigation often hinges less on inferring final targets---as conventionally done in collaborative manipulation---and more on social conventions for resolving encounters~\citep{Wolfinger95,mavrogiannis2022social}. Because such encounters are a form of joint action coordinated implicitly through motion~\citep{knepper_hri_2017}, early work argued that robots moving among people should shape their trajectories to be intent-expressive rather than merely goal-directed~\citep{kirsch,lichtenthaler:hal-01684307}. \citet{Fujioka2024} estimate human understanding online and adapt communicative signals encoded into robot motion online. \citet{mavrogiannis2018social} introduced Social Momentum (SM), in which the robot exaggerates its passing side to clearly signal its preferred avoidance strategy to co-navigating pedestrians. In a similar hallway-passing setting, \citet{cathcart2023opinion} use nonlinear opinion dynamics to let the robot rapidly and visibly commit to a passing side, breaking the deadlocks that arise when intent stays ambiguous. \citet{Bastarache} optimized trajectories that remain interpretable to nearby humans while ensuring predictability. \citet{taylor2022observer} and \citet{stefanos_visiblity} incorporate visibility constraints into legible robot motion planning, making sure that intent-expressive motion remains within a person’s field of view. Beyond expressing intent, \citet{geldenbott2024legible} couple legibility with an equitable sharing of collision-avoidance effort, linking how a robot signals its intent to how much of the avoidance burden it assumes. Legible SRN methods differ in how robot intent is represented and conveyed, ranging from destination-based goals to interaction-centric cues such as passing side. However, the comparative effects of these design choices on human motion and perception in embodied pedestrian encounters remain insufficiently understood.

\textbf{Hallway navigation}. Hallway navigation is ubiquitous,  motivating the incorporation of hallways into the design of user studies on SRN~\citep{Pacchierotti2005,Pacchierotti2006,smart2013}. Unlike open spaces, hallways introduce geometric constraints amplifying the need for agile, readable robot motion. Prior work in hallway navigation emphasized proxemics and side-passing conventions. \citet{Pacchierotti2005} proposed strategies where robots initiate avoidance early, shifting to the culturally appropriate side and yielding clearly to pedestrians. In follow-up work, they quantified human comfort for different passing distances, showing that distances that were too close invaded personal space, while excessively wide detours appeared inefficient~\citep{Pacchierotti2006}. \citet{smart2013} framed hallway navigation as a costmap-based problem where robots balance actions like yielding, slowing, or parking, and examined gaze as an additional cue. Inspired by these early investigations, our work introduces the feature of intent expressiveness into hallway navigation, analyzing when legibility cues matter, and how to generate them.

%% file: sections/technical.tex
\section{Foundations}

We mathematically formalize our problem domain of legible SRN and motivate our research questions.

\subsection{Domain}

\textbf{Social robot navigation}.
We consider a robot $r$ navigating in a workspace $\mathcal{W} \subseteq \mathbb{R}^2$, among static obstacles $\mathcal{O} \subset \mathcal{W}$ and $n \geq 1$ co-navigating humans. At each discrete time step $t$, the robot is in state $s^r_t$, and human $i$ is in state $s^i_t$. The robot is a dynamical system, evolving according to $s^r_{t+1} = f(s^r_t, u_t)$, where $u$ is a control action, drawn from a space of controls $\mathcal{U}$. Each human observes the robot and the other humans, and adapts their navigation according to an unknown policy $\pi_i$. The goal of the robot is to reach a destination $d\in\mathcal{W}$, while respecting the safety and comfort of co-navigating humans.

\textbf{Human inference in social robot navigation}. Studies on human action understanding suggest that humans tend to associate observed actions of others with possible goals in a given context~\citep{CSIBRA200760, BAKER2009329}. This can be observed in pedestrian domains as well, where humans engage in a form of negotiation over a socially acceptable way to resolve possible navigational conflicts~\citep{Wolfinger95}. For social navigation, we model this type of human inference as
\begin{equation}
    \mathcal{I}: \mathcal{S} \to \mathcal{G}\mbox{,}
\end{equation}
where $\mathcal{S}$ is a space of robot trajectories, and $\mathcal{G}$ represents a space of robot intentions. In practice, we assume that a human observes a partial robot trajectory $s^r_{a:b} = (s^r_a,\dots,s^r_b)$, from time $a$ to time $b$, and infers its intent as a belief $\mathcal{I}(s^r_{a:b})=P(g\mid s^r_{a:b})$, where $g\in\mathcal{G}$ is the robot's \emph{navigational intent}. While navigational intent is often modeled as the robot's intended destination $d$~\citep{kirsch,lichtenthaler:hal-01684307,taylor2022observer,Amirian2024,Dragan}, other options are possible (e.g., passing side, planned path, etc.). 

\textbf{Legible robot motion}. Following prior work~\citep{Dragan}, we refer to legibility as the property of robot motion that enables an observer to quickly and confidently infer the robot's intent. In a navigation setting within a shared context $M$, a robot may enable pedestrians to correctly infer its true intent $g^*$ through motion $s^r_{a:b}$, extracted by solving the following optimization problem:
\begin{equation}
    {s^r_{a:b}}^* \leftarrow \arg\max_{\mathcal{S}} \; P(g=g^* \mid s^r_{a:b}, M)\mbox{.}\label{eq:legibility}
\end{equation}
Here, by context $M$, we refer to any information that conditions this inference, such as the map, past robot and pedestrian behavior, or task-related cues, and it is where richer context would enter in a more general system. In our implementation, this inference relies only on the candidate intent set $\mathcal{G}$ and the cost $C$ of the observed trajectory.

Legibility is often discussed in relation with and contrasted from \emph{predictability}~\citep{Dragan}, the property of motion that matches what an observer expects of an agent heading toward a \emph{known} intent. Legible motion is the trajectory that maximizes the posterior, $\arg\max_{\mathcal{S}} P(g^* \mid s^r_{a:b})$ (eq.~\eqref{eq:legibility}), whereas predictable motion maximizes the likelihood, $\arg\max_{\mathcal{S}} P(s^r_{a:b} \mid g^*)$. By Bayes' rule, $P(g \mid s^r_{a:b}) \propto P(s^r_{a:b} \mid g)\,P(g)$, so the two coincide when efficient motion toward $g^*$ is unlikely under every other intent, $P(s^r_{a:b} \mid g) \approx 0$ for $g \neq g^*$, leaving $g^*$ as the most probable explanation of the observed motion. In navigation this can arise when intent is the destination. A corridor is one such case, where the efficient path toward one endpoint is incompatible with the other, so the predictable motion is also the most legible.

\textbf{Control for legible social robot navigation}. A flexible control paradigm that allows for the integration of legibility alongside standard considerations of safety and efficiency is model predictive control (MPC). MPC treats control as a receding-horizon optimization of the form:
\begin{equation}
\begin{split}
    u^*_{t:T} & = \arg\min_{u\in\mathcal{U}} \sum_{t}^{T-1}J(s^r_{t+1}, s^{1:n}_{t+1}) \\
    & s.t.\  s^r_{t+1} = f(s^r_t, u_t) \\
    & \qquad s^{1:n}_{t+1} = F(s^r_{t-h:t}, s^{1:n}_{t-h:t})\mbox{,} \label{eq:mpc}
\end{split}
\end{equation}
where $J$ is a cost function, $F$ is a human-motion model predicting future human motion given state history over a window $h$, and $T$ is a control horizon. We consider a composite cost $J=a_F J_F + a_L J_L$, comprising: a \emph{functional cost} ($J_F$) that captures human safety, efficiency, and obstacle avoidance as a weighted sum; a \emph{legibility cost} ($J_L$) that promotes early and confident communication of the robot’s intent. The legibility term motivates the selection of actions maximizing $P(g=g^* \mid s^r_{a:b}, M)$ from eq.~\eqref{eq:legibility} and is defined as
\begin{equation}
    J_L(s^r_{t+1}) = - w_t \, P(g = g^* \mid s^r_{t:t+1}, M)\mbox{,}
\end{equation}
where $w_t$ decays linearly over time, placing greater weight on earlier timesteps to encourage prompt intent disambiguation, consistent with prior work on legible motion~\citep{Dragan}. Intuitively, $J_L$ is low when the robot quickly biases an observer’s belief toward the correct intent, and large when the robot’s actions fail to provide early evidence of its goal.

\subsection{Modeling the Pedestrian's Belief}
\label{sec:legibility_framework}

Following prior work~\citep{Dragan}, we model a pedestrian's belief over the robot's intent by evaluating how well robot motion aligns with each candidate intent $g~\in~\mathcal{G}$, assuming the robot is optimizing a function $C(s^r_{a:b},g)$ representing the cost of reaching intent $g$ given the trajectory $s^r_{a:b}$. We express this belief as a probability:

\begin{equation}
P(g=g^* \mid s^r_{a:b}, M) = \frac{\exp\left(-\frac{1}{\lambda} C(s^r_{a:b}, g^*)\right)}{\sum_{g' \in \mathcal{G}} \exp\left(-\frac{1}{\lambda} C(s^r_{a:b}, g')\right)}\mbox{.}\label{eq:belief}
\end{equation}

\textbf{Pedestrian inattention}. The parameter $\lambda$, part of the robot's contextual understanding $M$, represents human inattention, which occurs when pedestrians are texting, conversing, carrying an object, or multitasking~\citep{Wickens2008,strayercell2001,Horrey2006}. This modeling decision is consistent with the concept of bounded rationality~\citep{gershman2015computational,griffiths2015rational}, which is often employed in human modeling across a range of domains~\citep{temp_bounded_rationality,Braun2015, Xu2024, Ortega2015}. Higher values of \(\lambda\) reflect reduced attention, resulting in diffuse, uncertain beliefs. When \(\lambda \to \infty\), the pedestrian is maximally distracted, unable to reason about robot intents, i.e., a uniform belief over all candidate goals:
    \begin{equation}
    \lim_{\lambda \to \infty} P(g \mid s^r_{a:b},M) = \frac{1}{|\mathcal{G}|}\mbox{.}
    \end{equation}
In contrast, lower $\lambda$ values yield confident, near-deterministic inferences. When \(\lambda \to 0^+\), the pedestrian’s inference is maximally confident that the intent of minimal cost is the true robot intent, i.e., the belief collapses to a point mass on the minimal-cost goal:
    \begin{equation}
            \lim_{\lambda \to 0^+} P(g \mid s^r_{a:b},M) =
    \begin{cases}
    1 & \text{if } g = \arg\min_{g' \in \mathcal{G}} C(s^r_{a:b}, g') \\
    0 & \text{otherwise}\mbox{.}
    \end{cases}
        \end{equation}

\subsection{Challenges in Generating Legible Motion for SRN}

Navigation domains in industrial, private, and public spaces feature rich context (e.g., geometric constraints, dense crowds, variable human internal state), motivating a deeper understanding of the role and relevance of legible robot motion. To this end, we approach the following research questions:

\textbf{Q1. How does intent representation shape legible SRN?}
    In the legibility framework of~\citet{Dragan}, intent refers to the robot's goal in the task space, e.g., the pose of a bottle that the robot intends to grasp. Much of the work on legible motion for SRN adopts a similar notion of intent: the robot aims to generate motion that is legible with respect to its destination~\citep{kirsch,lichtenthaler:hal-01684307,taylor2022observer,Amirian2024}. While this can be a practical assumption for simplified environments like sparsely populated open spaces, we expect that it does not scale to realistic environments with geometric constraints, rich contexts, and dense crowds.  Under such settings, pedestrians are not concerned with \emph{where} others are going but rather \emph{how} they could seamlessly avoid conflicts~\citep{Wolfinger95}, for example whether they would be passing each other on the left or the right side. Additionally, in earlier legibility work~\citep{kirsch,lichtenthaler:hal-01684307,taylor2022observer,Amirian2024}, the true intent $g^*$ is treated as fixed. For example, in the work of~\citet{Dragan}, when going for the right bottle, the robot needs to be consistently legible with respect to the right bottle.
    However, the dynamic nature of pedestrian domains may impact the validity of the assumption of fixed intent: pedestrians may modify their navigation intent to resolve conflicts with others. For instance, they might head toward an intermediate waypoint near their true destination.

\textbf{Q2. How does human distraction impact benefits of legible SRN?} The combination of functional and legible costs in eq.~\eqref{eq:mpc} highlights a fundamental tradeoff: legible motion can be at odds with functional efficiency~\citep{Dragan}. A variety of contextual factors could impact this tradeoff, including crowd density, spatial constraints, or human attentiveness. In this work, we focus on the latter. Prior work on legible motion in HRI assumes that the observer is fully engaged and attentive to robot behavior~\citep{Dragan,stefanos_visiblity,taylor2022observer,Legibilitydiffuser,wallkotter2022newapproachevaluatinglegibility}, a situation corresponding to small $\lambda$. In realistic pedestrian domains, however, human attention is often limited. Here, we want to understand how the legible SRN framework discussed in~Sec.~\ref{sec:legibility_framework} impacts navigation performance and human impressions when pedestrians around the robot are distracted, a situation expressed via larger $\lambda$ values. Under distraction, pedestrians may have limited capacity to register communicative signals broadcasted by the robot. We anticipate that in such cases, vividly legible motion cues that allow humans to rapidly disambiguate the robot’s immediate intent despite limited attention may reduce confusion and support safe conflict resolution. Approaching this question will help inform the design of SRN algorithms that adaptively balance efficiency and legibility.

%% file: sections/experiments.tex
\section{How does Intent Representation Shape Legible SRN}
 \label{sec:study1}

We present an IRB-approved lab study (HUM00268645) investigating how the intent representation underlying legible motion generation impacts navigation performance and user impressions (Q1). 

\subsection{Experiment Design}

\begin{figure}[t]
    \centering
    \includegraphics[width=0.95\linewidth]{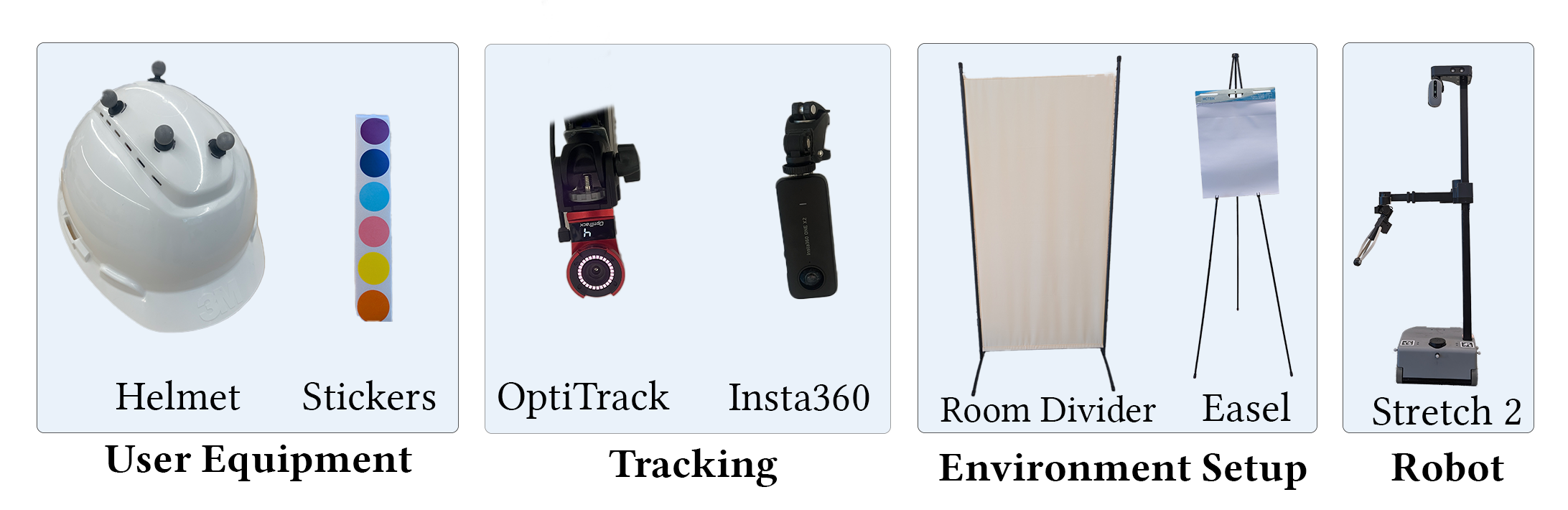}
    \Description{Four groups of labeled equipment photographs: user equipment (a white helmet fitted with motion-capture markers and a strip of six colored round stickers); tracking (an OptiTrack camera and an Insta360 camera); environment setup (a fabric room divider and an easel holding a paper pad); and the robot (the Stretch~2, a wheeled base with a tall vertical mast and a horizontal arm ending in a gripper).}
    \caption{Hardware and materials used across our studies. Participants wear a marker-equipped helmet tracked by the OptiTrack motion-capture system, while an overhead Insta360 camera records the interaction with consent. Room dividers form the walls of the hallway, and an easel pad at each end serves as a task station where the participant places the colored stickers. The robot is the Hello Robot Stretch~2. An action shot is shown in~Fig.~\ref{fig:experiment_setup}.}
    \label{fig:experiment_appartus}
\end{figure}

\textbf{Experimental setup}. The study took place in a hallway of size $2\,\mathrm{m} \times 5.8\,\mathrm{m}$, constructed with room dividers inside the lab (see~Fig.~\ref{fig:experiment_setup}). An easel pad was placed at each hallway end. We used the Hello Robot Stretch~2 mobile platform. An overhead camera (Insta360) recorded RGB video if the user gave consent. A motion capture system (OptiTrack) was used to localize the robot and the users. The full experimental apparatus is shown in Fig.~\ref{fig:experiment_appartus}. All navigation algorithms were implemented using a state-of-the-art MPC framework~\citep{williams}. Human motion prediction, $F$ in eq.~\eqref{eq:mpc}, was modeled as a constant velocity (CV) propagation, following extensive prior work~\citep{Kollmitz2015TimeDP,liu2023intentionawarerobotcrowd,singamaneni2021humanawarenavigationplannerdiverse,poddar2023crowdmotionpredictionrobot,scholler2020constantvelocitymodelteach}. The MPC horizon was set to 10 steps with a time step of $0.3\,\mathrm{s}$, and controllers operated at 20\,Hz. 

\textbf{Procedure}. Upon arriving in the lab, users provided informed consent, possibly opting out of video recording and data publishing. They were then given instructions, going over the task and the robot’s role, before participating in a trial round in which the robot remained stationary. They then completed five trials involving hallway navigation alongside the robot. Each trial lasted approximately 80 seconds. Right after each trial, users completed a questionnaire soliciting their reflections over their interactions with the robot. They then went through a debriefing form, provided demographic information and prior experience with robotics, and were compensated \$20. Each complete session lasted no more than 40 minutes.

\textbf{Task description}. Users were told that the study concerned human impressions of robot navigation algorithms but were not given its specific objectives. They were instructed to imagine that they were workers in a factory performing a mock “factory inspection” task. Completing an inspection task involved the user placing a colored sticker on an easel pad. To justify the robot's presence, users were told that it was monitoring their inspection and recording the sticker colors. In each study session, a user completed five trials, each corresponding to a different experimental condition, i.e., a different algorithm executed by the robot. In each trial, the user traversed the hallway back and forth six times to place six stickers across the two easel pads. Throughout the trial, the robot also traversed back and forth along the hallway, always moving against the user's direction, engaging in head-on encounters with them. Upon reaching its destination, the robot executed a brief turning maneuver to reorient itself toward the opposite end of the hallway.
To ensure repeatable human-robot encounters, we synchronized human and robot motion by instructing users to only turn around and move to the opposite easel after they hear a gong sound, signaling that the robot was ready to move.
Because the robot’s nominal speed ($0.33~m/s$) was lower than participants’ walking speeds, its navigation destinations were positioned closer than the human stations, ensuring that both agents reached their respective ends within comparable timeframes.

\subsection{Conditions}

Each participant completed five trials in a within-subjects design. In each trial, the robot executed a different navigation algorithm, with algorithm order counterbalanced using a Balanced Latin Square to mitigate ordering effects. All five strategies were implemented within the same model predictive control (MPC) framework, differing only in how the robot's intent is represented and conveyed. We evaluate the following strategies:

\textbf{Goal-based legibility (GL)}. Here, the space of intent, $\mathcal{G}$, is modeled as the set of the two hallway endpoints, with $g^*=d$, i.e., corresponding to the robot's intended destination, similar to prior legibility implementations by \citet{kirsch,Dragan}. The pedestrian's belief in eq.~\eqref{eq:belief} is computed using a Euclidean distance-based cost $C$. Intuitively, this algorithm is favoring actions that consistently communicate the robot's intended destination. Because both endpoints lie along the corridor axis, efficient motion toward the destination is also the legible motion, and GL thus corresponds to the case where legibility and predictability collapse.

\textbf{Passing-side legibility (PL).} Here, $\mathcal{G}$ consists of artificial subgoals placed to the left and right of the hallway endpoints ($d$), effectively creating the notion of passing-side interactions. By steering toward one of these offsets, the robot conveys its intended side of passage. The robot's true intent $g^*$ is assigned at random before the trial and held fixed for each encounter. The cost $C$ is identical to GL, and left/right assignments are balanced across trials for all participants.

\textbf{Dynamic passing side legibility (DPL)}. Here, $\mathcal{G}$ is defined as in PL, but $g^*$ is dynamically adapted at run time. Specifically, the robot selects the passing side associated with the lower predicted probability of being chosen by the human, based on CV predictions of the human’s motion. These probabilities are computed using the same Bayesian inference framework employed earlier for goal inference in legibility, i.e., $P(g \mid s^i_{0:t})$, but here applied to the human’s observed trajectory $s^i_{0:t}$ rather than the robot’s, to infer the human’s preferred passing side.

\textbf{Social Momentum (SM)}~\citep{mavrogiannis2022social}. Here, $\mathcal{G}=\{L, R\}$, where $L$ and $R$ represent passing from the left and right, respectively. Unlike PL and DPL, which encode passing-side intent through artificial offset goals, SM reasons about passing directly in the joint human–robot interaction space. This allows us to not fixate our results to the specific notion of generating passing sides using artificial goals. The cost $C$ is defined using the angular momentum of the system: its sign encodes the passing side, while its magnitude reflects confidence~\citep{mavrogiannis2022social}. Intuitively, $J_L$ rewards trajectories that reinforce the currently emerging passing side. Thus $g^*$ is dynamically adapted based on the passing preference of the human.

\textbf{No legibility (NL)}. This is vanilla MPC implemented as a non-legible reference algorithm, implemented by setting $J_L=0$. This baseline represents purely functional robot navigation without communicative intent.

\subsection{Implementation}

The controllers were implemented as ROS~2 nodes using PyTorch. We tuned each controller separately to balance goal progress and collision avoidance while producing its intended navigation behavior. Model predictive path integral (MPPI) control evaluated 250 sampled trajectories per update over a 10-step horizon with a timestep of $0.3\,\mathrm{s}$, and the control loop operated at 20\,Hz. Human velocity was estimated from consecutive tracked positions using a $0.05\,\mathrm{s}$ interval and used for constant-velocity prediction.

Table~\ref{tab:controller_parameters} reports the principal coefficients governing intent expression. For GL, PL, and DPL, the objective rewards a higher inferred probability of the selected intent and penalizes trajectory cost. The reward and regularization coefficients determine the contributions of these two terms to the MPC cost. At prediction step $k$ within the planning horizon, the probability reward is additionally weighted by $w_k$ to favor earlier intent communication. SM uses an angular-momentum reward instead of a probability-based reward, while NL omits the legibility term. Code for this work is available at \url{https://github.com/fluentrobotics/Legible_MPPI}.

\begin{table}[t]
    \centering
    \small
    \renewcommand{\arraystretch}{1.15}
    \caption{Selected parameters governing intent expression in the navigation controllers. Each controller was tuned separately to balance goal progress and collision avoidance while producing its intended navigation behavior.}
    \label{tab:controller_parameters}
    \Description{Four-row table with columns Parameter, Setting, and Role: the intent-probability reward coefficient (100 for GL, PL, and DPL), the trajectory regularization coefficient (0.1 for GL, PL, and DPL), the social-momentum reward coefficient (11 for SM), and the temporal legibility weight (decreasing from 10/15 to 1/15 over prediction steps 0 to 9 for GL, PL, and DPL), each with a short description of its role.}
    \begin{tabularx}{\linewidth}{@{}p{0.29\linewidth}p{0.25\linewidth}X@{}}
        \toprule
        \textbf{Parameter} &
        \textbf{Setting} &
        \textbf{Role} \\
        \midrule
        Intent-probability reward coefficient &
        $100$ for GL, PL, DPL &
        Rewards motion that increases the inferred probability of the
        selected intent \\

        Trajectory regularization coefficient &
        $0.1$ for GL, PL, DPL &
        Balances intent expression against trajectory cost \\

        Social-momentum reward coefficient &
        $11$ for SM &
        Rewards larger angular momentum when the predicted passing
        direction remains consistent \\

        Temporal legibility weight &
        $w_k=(10-k)/15$, $k=0,\ldots,9$, for GL, PL, DPL &
        Places greater emphasis on earlier predicted motion \\
        \bottomrule
    \end{tabularx}
\end{table}

\subsection{Measures}

Because human beliefs and inferences about robot intent are inherently latent and cannot be directly observed, we rely on established measures to evaluate how legible motion impacted human motion.

To objectively assess the impact of different algorithms on human behavior, we analyze human motion. We measure human path inefficiency using the Path Irregularity (Human PI) measure~\citep{Guzzi2013}, defined as the amount of unnecessary turning per unit path length $(\text{rad}/\text{m})$, and widely used in SRN research~\citep{christudy,mavrogiannis2018social,Gao2022,Tsoi2024,Biswas2022}. We also measure human motion abruptness using Human Average Acceleration (Human AA), defined as the average acceleration along a user’s full trajectory over a trial, yielding a single scalar value per algorithm per user. Human AA is widely used in SRN literature~\citep{christudy,Gao2022,mavrogiannis2018social,Biswas2022,Zhang2022} as a proxy for how much robot motion physically challenged a user: higher acceleration indicates more speed changes, indicating greater discomfort.

We also study user impressions collected via questionnaires. We use the \textit{Discomfort} and \emph{Competence} subscales of RoSAS~\citep{RoSAS}, presented in randomized order on 9-point Likert scales. We also use the \textit{Mental Demand, Physical Demand, Temporal Demand, Frustration, Performance}, and \emph{Effort} scales from NASA-TLX~\citep{hart1988development}, presented in a 21-point format. To assess perceived collision risk and intent clarity, we asked users to rate: (\textit{L1}) -- “The robot will bump into me in the future” (perceived collision risk); (\textit{L2}) -- “I was quickly and accurately able to tell where the robot wants to go” (perceived legibility, based on \citet{Dragan}), both presented as 7-point scales. Finally, we collected open-form responses to capture insights not covered by the structured scales.

\subsection{Hypotheses}

We study how different legibility implementations impact navigation performance and impressions by investigating the following:

\textbf{H1: ``Legible algorithms will be more positively perceived and enable higher user performance."}
We hypothesize that legible algorithms (GL, SM, PL, DPL) will lead to lower acceleration and more regular paths for the users, compared to non-legible algorithms (NL). We further expect legible algorithms to be rated as more competent and comfortable on the RoSAS scale, and to impose lower workload on users as measured by the NASA TLX, in contrast to non-legible algorithms.

\textbf{H2: ``Legibility over the robot’s \emph{passing side} will be more positively perceived and enable higher user performance compared to legibility over the robot’s goal."}
We hypothesize that passing side legibility (PL, DPL, SM) will enable faster and more accurate inference of the robot’s intent than goal-based legibility (GL), as reflected in survey responses. Passing side legibility is also expected to yield smoother trajectories (lower acceleration and more regular paths), and to be perceived as more competent, more comfortable, and less effortful for users compared to goal-based legibility.

\textbf{H3: ``Dynamically adapting the robot’s legibility intent based on user reaction will be more positively perceived and enable higher user performance compared to legibility over a fixed intent."}
We hypothesize that dynamic adaptation (SM, DPL) will outperform fixed intent legibility (PL) under the same passing side representation. Specifically, dynamic algorithms are expected to produce smoother human motion (lower acceleration and more regular paths), be perceived as more competent and comfortable, and require less effort and workload than fixed intent approaches.

\begin{figure*}[t]
    \centering
    \begin{subfigure}[t]{0.24\textwidth}
        \centering        \includegraphics[width=\textwidth]{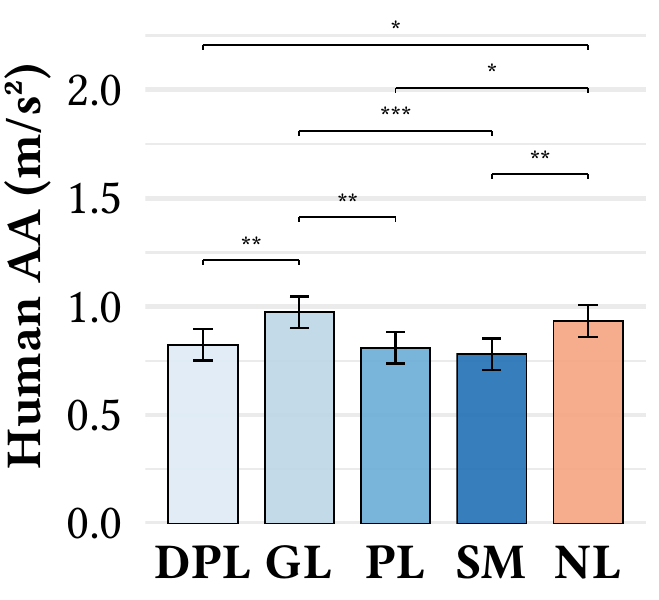}
        \subcaption[]{}
        \label{fig:avg_accel}
    \end{subfigure}\hspace{0.01\textwidth}\begin{subfigure}[t]{0.24\textwidth}
        \centering
        \includegraphics[width=\textwidth]{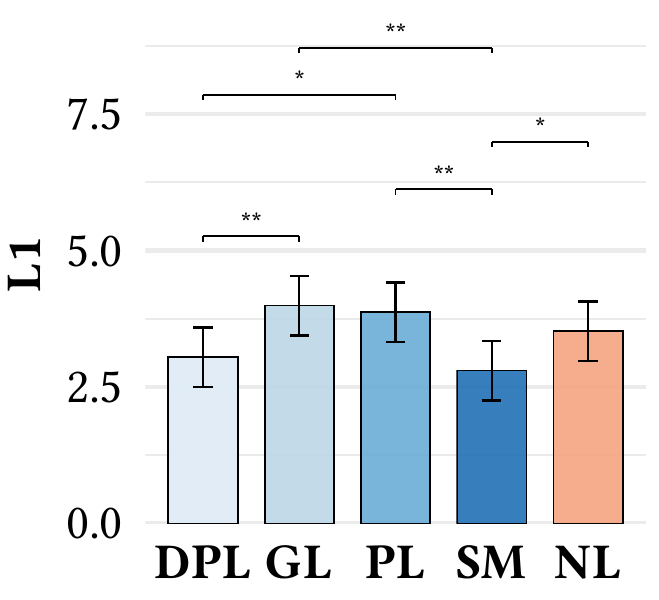}
        \subcaption[]{}
        \label{fig:L1}
    \end{subfigure}\hspace{0.01\textwidth}
    \begin{subfigure}[t]{0.24\textwidth}
        \centering
        \includegraphics[width=\textwidth]{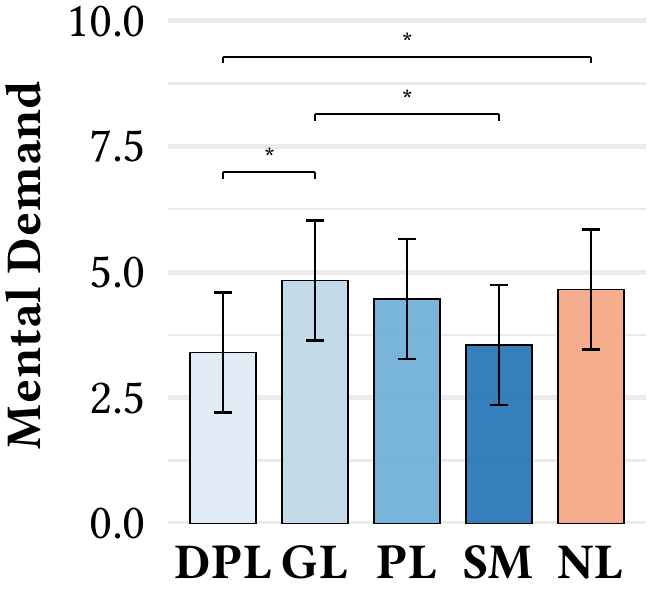}
        \subcaption[]{}
        \label{fig:mental_demand}
    \end{subfigure}\hspace{0.01\textwidth}\begin{subfigure}[t]{0.24\textwidth}
        \centering
        \includegraphics[width=\textwidth]{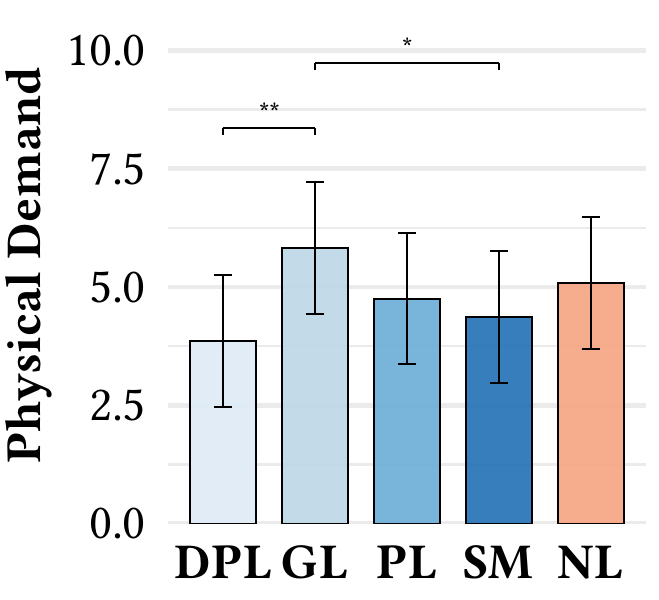}
        \subcaption[]{}
        \label{fig:physical_demand}
    \end{subfigure}
    \Description{Four bar charts, labeled (a) to (d), each showing estimated marginal means with 95\% confidence-interval error bars for the conditions DPL, GL, PL, SM, and NL, with the legible conditions in shades of blue and NL in orange: (a) human average acceleration in meters per second squared, (b) the L1 perceived collision-risk rating, (c) NASA-TLX mental demand, and (d) NASA-TLX physical demand. GL has the tallest bar in every panel, and brackets with asterisks mark significant pairwise differences, which include DPL versus GL and GL versus SM in all four panels.}
    \caption{Estimated marginal means (EMMs) with 95\% CI for Human Average Acceleration $(m/s^2)$, \textit{L1} (Likert Scale 1-7), Mental Demand (1-21), and Physical Demand (1-21). Significant pairwise differences, adjusted using the Benjamini--Hochberg (BH) method, are indicated by bars with asterisks: \textit{***}~$p < .001$, \textit{**}~$p < .01$,~\textit{*}~$p < .05$.}
    \label{fig:metrics}
\end{figure*}

\begin{figure*}[t]
    \centering
    \includegraphics[width = \linewidth]{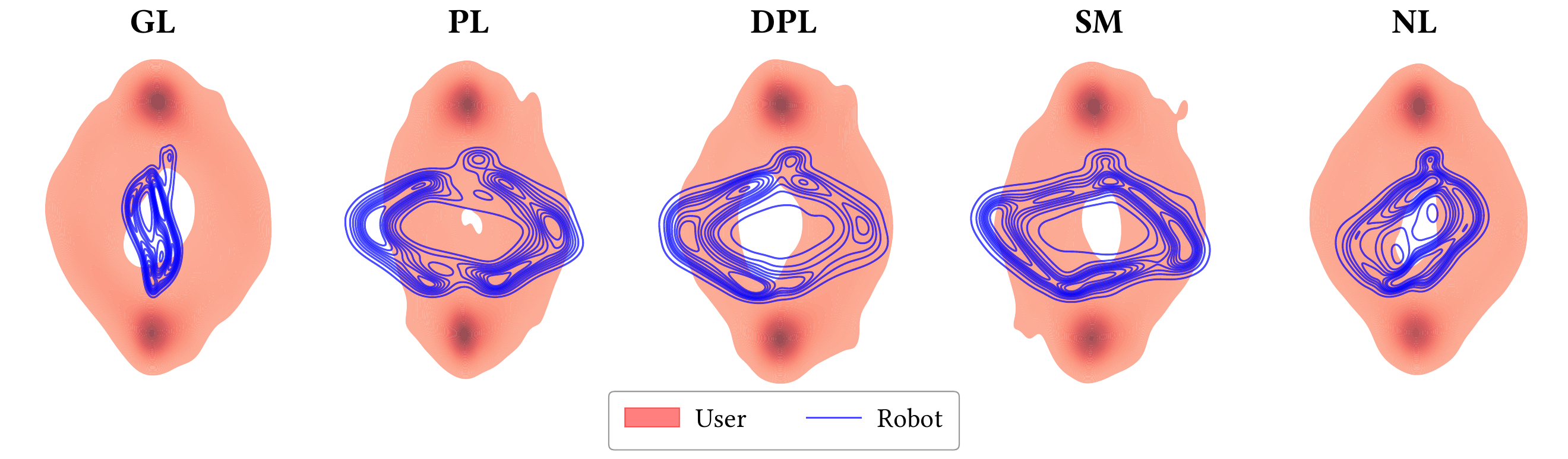}
    \Description{Five side-by-side top-down density maps, one per condition (GL, PL, DPL, SM, NL), each showing user positions as an orange shaded region, elongated vertically and darkest near its top and bottom ends, overlaid with blue contour lines for robot positions near the center. The robot contours form a narrow, nearly vertical band for GL; wide rings extending to both the left and right for PL, DPL, and SM; and a smaller ring for NL.}
    \caption{Trajectory density maps of the robot (blue contours) and users (orange heatmaps) under five motion strategies. GL produces narrow, endpoint-directed paths with no side preference. PL spreads robot density evenly across both sides, reflecting random passing choices. DPL shifts density adaptively opposite to the predominant human flow, showing context-aware side selection. SM generates the widest and earliest sideward deviations, using angular momentum to reinforce the emerging passing side. NL has narrower trajectories shaped only by collision avoidance.}
    \label{fig:trajectory_density}
\end{figure*}

\subsection{Analysis}

A total of 45 human subjects participated in the study, recruited from a university population through mailing lists. The participants (32 male, 12 female, 1 unidentified) had an average age of 22.42 years (SD = 2.75). On a 5-point Likert scale, they rated their familiarity with robotics technology at an average of 3.68 (SD = 1.03). Qualitative differences across algorithms are shown in~Fig.~\ref{fig:trajectory_density}.

We modeled the study metrics using linear mixed-effects regression, with algorithms and algorithm order as fixed effects, and random effects for participants to account for repeated measures. Pairwise comparisons between algorithms were corrected using the Benjamini–Hochberg (BH) method.

We present the results organized by hypothesis and grouped by metric type. Table~\ref{tab:competence_discomfort} reports subjective RoSAS ratings of \textit{Competence} and \textit{Discomfort} whereas Fig.~\ref{fig:metrics} shows objective and workload-related outcomes, including \textit{L1} (Fig.~\ref{fig:metrics}\subref{fig:L1}), mental demand (Fig.~\ref{fig:metrics}\subref{fig:mental_demand}), physical demand (Fig.~\ref{fig:metrics}\subref{fig:physical_demand}), and Human AA (Fig.~\ref{fig:metrics}\subref{fig:avg_accel}). No statistically significant effects were observed for Human PI, \textit{L2}, or the NASA-TLX subscales of Performance, Effort, and Temporal Demand.

\begin{table}[ht]
\centering
\caption{Estimated marginal means (EMMs, 95\% CI) of RoSAS Competence and Discomfort ratings for navigation algorithms.
Different letters indicate significant pairwise differences ($p < .05$), with comparisons adjusted using the Benjamini–Hochberg (BH) method. No letters indicate no significant pairwise comparisons. Overall, the adaptive passing-side strategies (SM, DPL) were rated most competent, and DPL also elicited the lowest discomfort, whereas goal-based legibility (GL) was rated worst on both.}
\label{tab:competence_discomfort}
\Description{Table with one row per algorithm (SM, DPL, PL, NL, GL) giving the mean and 95\% confidence interval of RoSAS competence (higher is better) and discomfort (lower is better), with superscript letters marking significant differences. SM and DPL have the highest competence (5.75 and 5.66), DPL the lowest discomfort (2.30), and GL the lowest competence (4.62) and highest discomfort (3.09).}
\begin{tabular}{lcc}
\toprule
\textbf{Algorithm} & \textbf{Competence $\uparrow$} & \textbf{Discomfort $\downarrow$} \\
\midrule
SM   & 5.75 [5.19, 6.31]$^{a}$ & 2.52 [2.04, 2.99] \\
DPL  & 5.66 [5.11, 6.22]$^{a}$ & 2.30 [1.83, 2.78]$^{b}$ \\
PL   & 5.05 [4.49, 5.61]$^{b}$ & 2.95 [2.48, 3.43]$^{a}$ \\
NL   & 5.05 [4.50, 5.61]$^{b}$ & 2.53 [2.06, 3.01] \\
GL   & 4.62 [4.06, 5.17]$^{b}$ & 3.09 [2.62, 3.57]$^{a}$ \\
\bottomrule
\end{tabular}
\end{table}

\textbf{H1}. NL was perceived as significantly less competent than SM and DPL ($p < .05$), and more likely to bump into participants than SM (\textit{L1}, $p < .05$). No significant differences emerged between NL and PL or GL in terms of subjective metrics. NL also elicited higher mental demand and frustration compared to DPL ($p < .05$). Qualitative responses reinforced these impressions: participants described NL as \emph{“hard to predict,”} noting that the robot was \emph{“almost always in my way”} or \emph{“super slow to respond”}. By contrast, DPL was praised for \emph{“giving me space,”} and SM for \emph{“interacting at the right time”}. These remarks align with the higher competence ratings for SM and DPL.

Objective metrics corroborated the subjective findings. NL produced significantly higher Human AA than DPL and PL ($p < .05$) as well as SM ($p < .01$), indicating more abrupt changes in human motion under NL.

Taken together, these results indicate that legible robot motion has positive effects, but its impact depends on how intent is represented. Passing-side formulations with dynamic adaptation (DPL, SM) received higher competence ratings and elicited lower human average acceleration compared to the non-legible baseline (NL). However, the absence of consistent subjective differences for PL and GL, despite PL showing objective improvement in Human AA, suggests that not all forms of legibility are equally effective. Thus, \textbf{H1} is only partially supported.

\textbf{H2}. GL was consistently rated as the least competent among the legible conditions. Subjectively, participants rated GL as more likely to collide (\textit{L1}, $p < .01$) and less competent than SM and DPL ($p < .001$), and more discomforting than DPL ($p < .05$). GL was also associated with significantly higher mental demand, frustration and physical demand compared to DPL and SM ($p < .05$). Open ended responses echoed this: participants frequently mentioned that GL \emph{“ignored my location,”} \emph{“took the same path I was moving,”} and made them feel like they were \emph{“about to bump into the robot each time.” } These remarks directly capture the higher discomfort observed in the quantitative measures.

Objective measures mirrored these impressions. GL yielded the highest Human AA, reflecting more abrupt, less smooth human motion, whereas DPL, PL, and SM maintained significantly lower values ($p < .01$), supporting smoother interactions.

These findings suggest that, in the hallway setting evaluated here, destination-based legibility (GL), as adopted in prior work~\citep{kirsch}, yields less smooth human motion and higher workload than passing-side formulations with dynamic adaptation (SM, DPL). While PL and GL differed in terms of Human AA, subjective differences emerged only between the adaptive passing-side strategies (DPL, SM) and GL. This indicates that adaptation amplified the perceived benefits of passing-side legible motion. Thus, \textbf{H2} is only partially supported.

\textbf{H3}. PL was rated as significantly less competent than SM and DPL ($p < .05$) and more likely to collide (\textit{L1}, $p<.05$). It also elicited higher discomfort than DPL ($p<.05$). Open-ended responses reinforced these ratings: while some participants acknowledged that PL \emph{“chose a path that wouldn’t collide,”} others remarked that it \emph{“didn’t adapt to my walking style”} or felt \emph{“aimless.”} These comments capture why, despite passing-side intent representation, PL was still perceived as less competent and more uncomfortable.

Objective metrics provided a more mixed picture. PL showed slightly lower Human AA than DPL but higher than SM, though these differences were not statistically significant.

Taken together, the evidence indicates that given the passing-side intent representation, dynamic adaptation improves perceptions of competence and comfort, although objective measures did not show significant differences. Thus, \textbf{H3} is partially supported.

\textbf{Summary.} Across all three hypotheses (\textbf{H1–H3}), results converge on a consistent pattern: legible motion improves both perceived competence and the smoothness of human motion, but the effectiveness depends on how intent is represented. Passing-side formulations with dynamic adaptation (DPL, SM) produced the most favorable outcomes, consistently outperforming the baseline (NL) and, in most cases, the static passing-side strategy (PL). In contrast, goal-based legibility (GL) elicited higher workload and less fluent trajectories, making it less suitable for close human–robot navigation. Taken together, the findings indicate that interaction-level representations of intent, combined with dynamic adaptation, are most suitable for legible robot behavior in hallway navigation.

\subsection{Exploratory Analysis}
\label{sec:study1_legibility_objective}

In the foundational legibility study of \citet{Dragan}, legibility was assessed with passive
observers. Participants watched recorded reaching motions and, from only the first portion of each
trajectory, predicted the robot's intended goal and rated their confidence, with motion deemed more
legible when it elicited earlier and more confident correct predictions. This protocol fits
manipulation, where the human only watches, but not our setting, where participants are
simultaneously observers and actors who perceive the robot and adjust their own motion in response,
and we want to capture their experience of legibility as co-navigators. Our self-report item (L2: ``I was quickly and accurately able to tell where the robot wants to go'') is the questionnaire analog of this goal-prediction measure, yet it showed no significant differences across conditions. We therefore conducted a post hoc exploratory analysis using two behavioral proxies that recover the same notion of early, confident intent disambiguation from the recorded interaction itself, computed per head-on encounter from the human and robot trajectories and averaged within each trial:

\begin{itemize}

    \item \emph{Conflict-resolution time}. \citet{olivier2012minimal} introduce the Minimal Predicted Distance
    (MPD), the closest separation the robot and human would reach if both held their current velocity, and
    \citet{vassallo2017how} track its evolution over an interaction to characterize how a crossing is resolved. For a
    head-on encounter with human $i$, let $\Delta p_t = p^r_t - p^i_t$ and $\Delta v_t = v^r_t - v^i_t$ be the
    relative position and velocity of the robot and human (from states $s^r_t$ and $s^i_t$) at step $t$. The
    predicted time to closest approach is
    $\tau_t = \max\!\big(0,\,-\langle \Delta p_t, \Delta v_t\rangle / \lVert \Delta v_t\rVert^{2}\big)$, giving
    $\mathrm{MPD}_t = \lVert \Delta p_t + \tau_t\,\Delta v_t\rVert$. We report the conflict-resolution time
    $t_c - t^{\ast}$, where $t_c$ is the closest encounter and
    $t^{\ast} = \min\{\, t \le t_c : \mathrm{MPD}_{t'} \ge d_c \ \forall\, t' \in [t, t_c]\,\}$ is the earliest step
    from which the predicted clearance stays above a comfort threshold $d_c$. A larger value indicates that the human
    disambiguated the robot's passing intent earlier. We set $d_c = 0.8\,\mathrm{m}$, a comfortable center-to-center passing distance~\citep{neggers2022speed}, and confirmed that the ranking remained stable for $d_c \in [0.6, 1.0]\,\mathrm{m}$.

    \item \emph{Robot responsibility} ($R_\text{robot}$)~\citep{probst2025responsibility}. We quantify the
    conflict at each step as $C_t = \max\!\big(0,\,1 - \mathrm{MPD}_t/d_0\big)$, which grows as the predicted
    clearance drops below the combined agent radius $d_0$. Each step-to-step change in $C_t$ is split between the two
    agents by a counterfactual comparison: the robot's contribution is how much the conflict changes due to its
    velocity adjustment, measured against it keeping its previous velocity, and analogously for the human.
    $R_\text{robot} \in [0,1]$ is then the robot's share of the total conflict reduction, so higher values indicate
    that the robot performed more of the avoidance.

\end{itemize}

We analyze both metrics with the same linear mixed-effects model used for our other measures. Both differed
significantly across conditions (Table~\ref{tab:legibility_objective}). Conflict-resolution time was longest
under the passing-side strategies (SM, DPL, PL), indicating that humans committed to a passing side earlier, and
shortest under goal-based legibility (GL) and the non-legible baseline (NL), where comfortable clearance emerged
only just before the closest encounter and the passing intent stayed ambiguous until late in the approach. Robot
responsibility followed the same ordering, with the robot performing the largest share of the avoidance under SM
and DPL and the smallest under GL.

\begin{table}[t]
\centering
\caption{Estimated marginal means (EMMs, 95\% CI) of conflict-resolution time and robot responsibility
($R_\text{robot}$) by algorithm in Study~1. Different superscript letters within a column indicate significant pairwise
differences ($p<.05$), adjusted using the Benjamini--Hochberg (BH) method. The passing-side strategies
(SM, DPL, PL) resolve conflicts markedly earlier and take a larger share of the avoidance than goal-based
legibility (GL) and the non-legible baseline (NL).}
\label{tab:legibility_objective}
\Description{Table with one row per algorithm (SM, DPL, PL, NL, GL) giving the mean and 95\% confidence interval of conflict-resolution time in seconds and of robot responsibility, both higher is better, with superscript letters marking significant differences. SM has the highest values on both (1.63 seconds and 0.23), DPL and PL are intermediate, and NL and GL are lowest on both.}
\begin{tabular}{lcc}
\toprule
\textbf{Algorithm} & \textbf{Conflict-resolution time (s) $\uparrow$} & \textbf{$R_\text{robot}$ $\uparrow$} \\
\midrule
SM   & $\mathbf{1.63}\,[1.29, 1.96]^{a}$  & $\mathbf{0.23}\,[0.20, 0.25]^{a}$ \\
DPL  & $1.16\,[0.82, 1.50]^{b}$  & $0.19\,[0.16, 0.22]^{b}$ \\
PL   & $1.22\,[0.88, 1.56]^{b}$  & $0.18\,[0.15, 0.20]^{b}$ \\
NL   & $0.28\,[-0.05, 0.62]^{c}$ & $0.14\,[0.12, 0.17]^{c}$ \\
GL   & $0.08\,[-0.26, 0.41]^{c}$ & $0.09\,[0.06, 0.12]^{d}$ \\
\bottomrule
\end{tabular}
\end{table}

These behavioral differences mirror our subjective and motion results: the adaptive passing-side strategies
(SM, DPL) that resolved conflicts earliest and bore the most avoidance were also rated more competent and yielded
smoother human motion, whereas goal-based legibility behaved much like the non-legible baseline. The intended
differences in legibility were thus expressed in coordination behavior, reinforcing that the benefit of legible
motion depends on how intent is represented and adapted to the human.

\section{How Human Distraction Impacts Benefits of Legible SRN}
\label{sec:study2}

In our IRB-approved Study 2, we investigate the role of legible motion under distraction (Q2). The distraction factor was implemented as a between-subjects manipulation: the no-distraction data were directly reused from Study 1, and a new cohort of participants was assigned to the distraction condition in Study 2.

\begin{figure}[t]
    \centering
    \includegraphics[width = \linewidth]{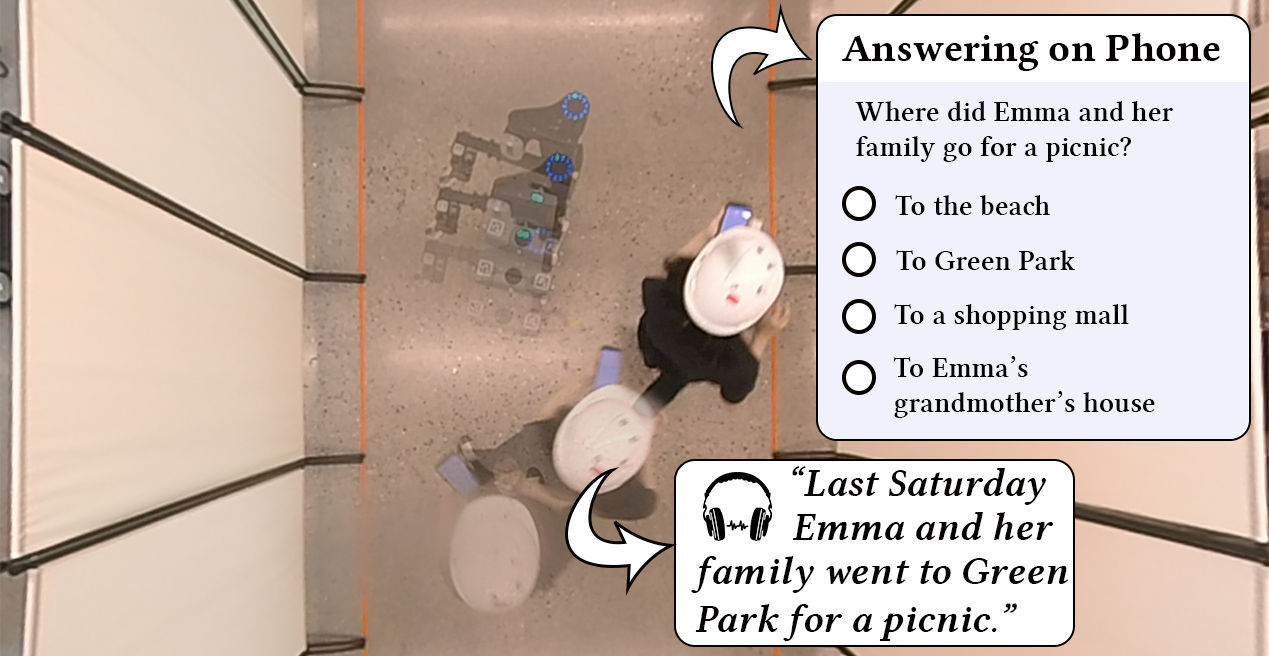}
    \Description{Overhead photograph of the hallway showing a participant in a white helmet holding a phone, with faded copies of the participant and the robot marking their successive positions as they pass each other. Two callouts illustrate the distraction task: a headphone icon beside an audio excerpt, ``Last Saturday Emma and her family went to Green Park for a picnic,'' and a phone question, ``Where did Emma and her family go for a picnic?'', with four answer choices.}
    \caption{Experimental setup illustrating the distraction condition. A participant listens to an audio passage through earbuds and answers comprehension questions on a phone while walking in the hallway. Divided attention leads the participant to veer closer to the robot, perceive a potential collision, and adjust their path.}
    \label{fig:distraction_setup}
\end{figure}

\subsection{Study Design}

We used the same task setup as Study 1, where users performed the mock inspection task of placing colored stickers on easels while navigating in a hallway alongside the robot.

\textbf{Distraction task}. Besides navigating, users simultaneously completed a verbal comprehension task: listening to an audio passage and answering related multiple-choice questions on a mobile device (see~Fig.~\ref{fig:distraction_setup}). This form of distraction was motivated by prior research on distracted pedestrian walking behaviors~\citep{pedestrian_distraction_gait_study,DistractionFromSmartphonesChangedPedestriansWalkingBehaviorsinOpenAreas}. The audio materials were modeled after standardized English comprehension passages (e.g., TOEFL listening sections) to provide a consistent cognitive challenge across users. Each passage was narrated at a rate of 90 words per minute and at a fixed volume level. In a pilot study with five participants, all were able to follow navigation instructions while answering the comprehension questions and reported the dual-task setup as manageable, indicating that the distraction was effective without being overwhelming. 

\textbf{Procedure}. Users completed consent forms, read instructions, and participated in a practice round. They then completed three trials involving hallway navigation alongside the robot, while concurrently completing the distraction task. Each trial lasted approximately 80 seconds. Right after each trial, users completed a questionnaire with our subjective measures. They then went through a debriefing form and were compensated \$15. The full session lasted up to 30 minutes. 

\textbf{Conditions}. We used a within-subjects design where in each trial the robot ran a different algorithm: Dynamic Passing-side Legibility (DPL), Social Momentum (SM), and No Legibility (NL). DPL and SM were chosen as the top legible performers from Study 1, whereas NL serves as a non-legible reference.
Similar to Study 1, condition order was counter-balanced using a Balanced Latin Square.

\textbf{Hypothesis H4: ``Legible algorithms will be more positively perceived and enable higher user performance in the presence of cognitive distraction, compared to respective ratings and performance under no distraction."} We expect that distraction will make the navigation task more challenging to complete, making legible motion essential for conflict resolution.
When users are solely focused on navigation, they may split the responsibility for collision avoidance with the robot. In contrast, when their attention is divided, we expect the navigation task to become more challenging, and anticipate that even subtle communicative signals on the robot's part will facilitate navigation.

\subsection{Analysis}

A total of 45 participants (28 male, 17 female; $M_{age}=23.55$, $SD_{age}=4.05$) took part in Study 2. On average, they reported robotics familiarity of 3.40 ($SD=0.91$) on a 5-point Likert scale. To validate the distraction manipulation, we compared self-reported mental demand and walking metrics across conditions. Participants reported significantly higher mental demand under distraction ($8.58 \pm 0.72$) than no-distraction ($4.14 \pm 0.60$), confirming that the manipulation increased cognitive load ($t(241.9) = -8.87,p < .001$). Task performance further supported this, with 91.1\% of participants scoring perfectly. Robot acceleration was higher under distraction ($1.70 \pm 0.51~m/s^2$ vs.\ $1.34 \pm 0.21~m/s^2$), indicating more abrupt adjustments when participants were less attentive. Finally, distraction reduced average walking velocity ($0.960 \pm 0.02$ vs.\ $1.11 \pm 0.01$, $t(244.23) = 8.71; p < .001$), further supporting the validity of the manipulation.

\begin{figure}[t]
    \centering
    \includegraphics[width=0.3\linewidth]{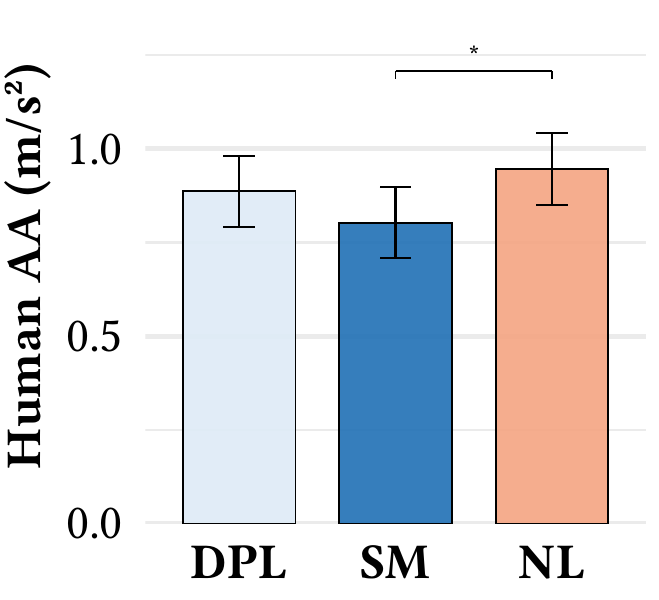}
    \Description{Bar chart of estimated marginal means of human average acceleration with 95\% confidence-interval error bars for DPL (about 0.89), SM (about 0.80), and NL (about 0.95), with DPL and SM in shades of blue and NL in orange. A bracket with one asterisk marks the significant difference between SM and NL.}
    \caption{Estimated marginal means (EMMs) with 95\% CI for Human AA $(m/s^2)$ by algorithm in Study~2. Significant pairwise differences are indicated by bars with asterisks: \textit{*} $p < .05$.}
    \label{fig:acceleration_emm}
\end{figure}

We compare algorithms under distraction using a linear mixed-effects regression model with the same specification as Study 1. We did not find significant differences between algorithms under distraction in terms of subjective measures. However, Human AA was lower when the robot was running legible algorithms (DPL, SM) than for NL (Fig.~\ref{fig:acceleration_emm}), with the NL-SM difference reaching statistical significance ($p < .05$). The reduction in Human AA from legible algorithms relative to the non-legible baseline was similar across distraction and no-distraction conditions (drawn from Study 1, with the comparison restricted to the algorithms used in both studies), incorporating repeated measures through participant random intercepts.

\textbf{H4}. We test H4 by fitting linear mixed-effects models with fixed effects of Algorithm, Distraction level, and their interaction, and random intercepts for participants. Likelihood ratio tests comparing models with and without the interaction term showed no evidence that distraction altered the performance of any algorithm for any of the metrics. Thus, \textbf{H4 was not supported}: legible algorithms did not enable significantly higher performance and impressions under distraction compared to the non-distraction setting, based on models restricted to algorithms present in both conditions.

\subsection{Exploratory Analysis}

We explore our collected dataset to uncover a deeper understanding of how distraction impacted the effects of legible motion for users.

\textbf{Distraction reduced variability in subjective ratings}. Residual variance in competence and discomfort ratings was lower under distraction (competence: $1.24$ vs.\ $1.01$; discomfort: $0.71$ vs.\ $0.46$). AIC and likelihood ratio tests favored a homoskedastic over heteroskedastic model ($\Delta$AIC = 1.5; $\chi^2(2) = 2.47; p = .29$), indicating that variance was reduced \emph{overall} rather than driven by specific algorithms. This indicates that once participant, baseline, and order effects were controlled for, algorithms explained little additional variance in competence or discomfort ratings. In practice, participants under distraction seemed to provide more uniform evaluations (e.g., “the robot was fine/safe enough”), making it harder to detect algorithm-level differences.

\textbf{Legible motion improved Human AA under distraction}. Human AA residual variance increased from $0.041$ (no-distraction) to $0.072$ (distraction), reflecting more diverse walking under distraction. We also found that users walked with significantly lower acceleration next to SM, compared to the non-legible baseline NL ($p < .05$). This suggests that legible motion continued to promote smoother coordination, although this was not registered in users' subjective ratings.

\textbf{Coordination benefits of legible motion persisted under distraction}. We repeated the exploratory analysis of Sec.~\ref{sec:study1_legibility_objective} on the distraction data, and the Study~1 pattern held. Conflict-resolution time was longest under SM ($2.35$ s), shorter under DPL ($1.78$ s), and shortest under the non-legible baseline NL ($0.81$ s), with all pairwise differences significant ($p<.01$). Robot responsibility was higher under both legible algorithms (SM, $R_\text{robot}=0.24$; DPL, $0.23$) than under NL ($0.20$; $p<.01$), although SM and DPL did not differ. These rankings reproduce Study~1, where the same strategies resolved conflicts earlier and took on more of the avoidance than NL. Crucially, these coordination benefits held even though subjective ratings showed no algorithm-level differences under distraction. Together with the lower Human AA for legible algorithms, this suggests that legible motion continued to shape how conflicts were resolved even when its effects were less apparent in participants' subjective impressions.

\begin{figure}[t]
\centering
\includegraphics[width=\linewidth]{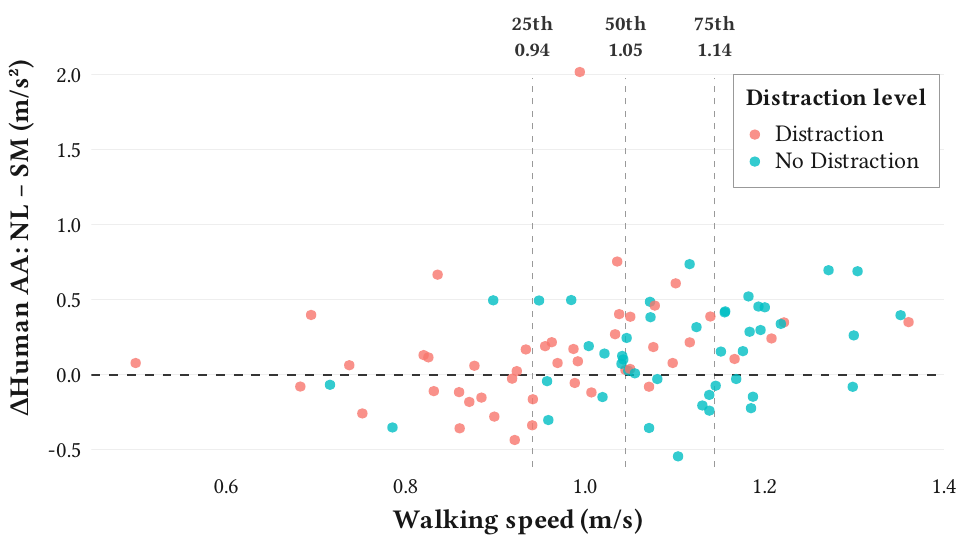}
\Description{Scatter plot of walking speed in meters per second (horizontal axis, about 0.5 to 1.4) against the NL minus SM difference in human average acceleration (vertical axis, about -0.5 to 2), with red points for distraction and teal points for no distraction, a dashed horizontal line at zero, and dashed vertical lines at the 25th, 50th, and 75th speed percentiles (0.94, 1.05, and 1.14). Distraction points lie mostly at slower speeds than no-distraction points, and most points fall between about -0.5 and 0.75, with one distraction point near 2.}
\caption{Scatter plot showing participant-level differences in Human AA (NL--SM) against mean walking speed under distraction and no distraction. Each point represents one participant. Positive values indicate lower Human AA and less abrupt velocity changes under SM. Dashed vertical lines mark the pooled 25th, 50th, and 75th walking-speed percentiles used in the exploratory analysis. SM improved motion smoothness over NL in both distraction and no-distraction conditions ($p < .05$), though the SM--NL difference did not vary significantly between distraction levels.}
\label{fig:distraction_nodistraction}
\end{figure}

\textbf{Legible motion benefits were more pronounced for higher human walking speeds}. The scatter plot in~Fig.~\ref{fig:distraction_nodistraction} illustrates the shift toward slower walking under distraction and the variation in the NL--SM Human AA difference across walking speeds in both attention conditions. Users walked significantly more slowly under distraction ($p < .001$). The acceleration difference was more often positive, suggesting that humans walked more unobstructed next to a legible robot, even in the distraction condition, with the effect more pronounced at higher speeds. We modeled speed as a continuous moderator with fixed effects of Algorithm, Distraction level, Speed, and their interactions, plus random intercepts for participants. The three-way interaction was not significant, but planned contrasts at the 25th, 50th, and 75th percentiles of pooled velocities showed stronger effects at higher speeds. Under distraction, SM consistently outperformed NL, with the gap widening as pace increased ($\Delta = 0.12, p = .020$; $\Delta = 0.19, p < .001$; $\Delta = 0.25, p = .001$). In the no-distraction condition, effects were not significant at slower or medium speeds, likely because fewer participants walked at these levels, but became significant at faster speeds ($\Delta = 0.17, p = .001$), though still smaller than under distraction (0.17 vs.\ 0.25).

\textbf{Summary}. Our analysis suggests that legible motion can be effective under distraction in hallway navigation. Although the overall interaction for speed was not significant, planned contrasts showed a widening NL–SM gap at higher speeds, indicating that benefits became more pronounced under higher navigation demands. In particular, subjective ratings were less sensitive under distraction, as participants judged the robot more uniformly, but objective measures suggested that legible motion, particularly SM, led to smoother human motion. The persistence of these benefits motivates future work on exploring the implications of different $\lambda$ values (eq.~\eqref{eq:belief}) and the development of systems that automatically adapt $\lambda$ to their surrounding context to strike an efficient balance between functional and legible behavior.

%% file: sections/discussion.tex
\section{Discussion}

Across two studies, we approached the question of how we should adapt the legible motion framework to SRN. Our findings indicate that the effectiveness of legible motion depends critically on how intent is represented and on the attentional state of human partners. Together, these results refine how legibility should be conceptualized and operationalized in social navigation settings. All algorithms were implemented within a common, carefully engineered model predictive control framework, enabling comparisons of different intent representation and adaptability approaches. Below we discuss the implications of these findings for legible motion generation in hallway settings and design of social navigation systems, as well as limitations and directions for future work.

\subsection{Intent representation shapes legible motion}

A central insight from this work is that legibility in social robot navigation should not be treated as a straightforward extension of destination-based formulations developed for isolated, goal-directed tasks. In the hallway setting tested in Study~1, the destination-based formulation adopted in prior work~\citep{kirsch,lichtenthaler:hal-01684307,taylor2022observer,Amirian2024,Dragan} was associated with higher reported workload and less smooth human motion, whereas the adaptive passing-side strategies supported smoother coordination and received higher competence ratings.

Our exploratory objective analysis reinforces this picture from a behavioral standpoint. Compared with goal-based legibility, passing-side strategies, and the adaptive variants in particular, enabled pedestrians to commit to a passing side earlier and led the robot to take on a larger share of the avoidance. Because these measures reflect how the encounter is jointly resolved rather than how it is perceived, their agreement with the subjective ratings indicates that interaction-level legibility shapes coordination behavior, not only impressions.

\subsection{Adaptation strengthens legible motion}

Beyond intent representation, our results highlight the importance of adaptability in legible motion. While static passing-side legibility improved objective smoothness relative to a non-legible baseline, its benefits were weaker and less consistently reflected in subjective measures compared to dynamically adaptive strategies. Open-ended responses suggest that fixed strategies were sometimes perceived as rigid or unresponsive, even when they avoided collision. In contrast, adaptive strategies continuously adjusted their expressed intent in response to human motion, reinforcing the reciprocal nature of navigation. This pattern highlights the value of using pedestrian motion to guide how the robot selects and updates its passing intent, strengthening the perceived benefits of clear passing-side signaling. These findings support treating pedestrians as interactive agents whose movements and emerging passing preferences inform the robot's behavior throughout the encounter~\citep{trautmanijrr,mavrogiannis2022social}.

\subsection{Impact of social norms}

Our design partly disentangles legibility from adherence to social norms. Passing-side legibility (PL) signals a side clearly but assigns it at random, so its coordination gains over the non-legible (NL) and goal-based (GL) baselines stem from early signaling rather than just from choosing a norm-appropriate side. A passing-side analysis is consistent with this. A side convention was clearly at play (69\% of encounters resolved to the same side, $p<.001$), but adherence to it did not track performance. Because GL and NL do not signal a side, it was set by the human, so they most often resolved on the conventional side (81\% and 76\%) yet were rated lowest. PL, in contrast, deviated from the convention most (51\%) while still achieving better objective coordination than both baselines. However, the discomfort voiced with NL and GL, described as ``almost always in my way'' or taking ``the same path I was moving,'' may also reflect perceived obstruction, path overlap, or encroachment on personal space. These concerns intersect with legibility because early communication of the robot’s near-term motion can help pedestrians anticipate whether their path will be obstructed and where they can pass. For the adaptive strategies (DPL, SM) especially, legible signaling and socially appropriate accommodation remain intertwined by construction. More fundamentally, social norms such as personal-space expectations and passing-side conventions are intrinsic to social navigation and interact with legibility~\citep{principles}. In this setting, motion that clarifies the robot’s intent may also resolve path conflicts, reduce perceived collision risk, and accommodate the pedestrian’s preferred route, making their effects difficult to separate.

\subsection{Benefits persist under divided attention}

Study~2 examined how human attention modulates the impact of legible motion by introducing a realistic distraction task. Under distraction, subjective ratings of competence and discomfort became less sensitive to algorithmic differences, yet objective measures of coordination, such as Human AA, the time taken to resolve conflicts, and the robot's share of the avoidance, continued to favor legible strategies over a non-legible baseline. This dissociation suggests that legible motion can influence interaction dynamics even when pedestrians are absorbed in secondary tasks. One interpretation is that under divided attention, pedestrians rely more heavily on peripheral, low-level motion cues than on explicit inference. In such cases, legibility might operate at a behavioral level, supporting smoother coordination without necessarily registering as a distinct qualitative experience. Exploratory analyses further indicate that the benefits of legible motion under distraction were more pronounced at typical and higher walking speeds, where coordination demands are greater. Comparing the two studies also suggests that attention may influence the subjective benefits of legibility. In Study~1, when navigation was the sole task, the adaptive strategies DPL and SM were rated more competent than NL, whereas Study~2 found no significant subjective differences between algorithms under distraction. Since pedestrians must perceive the robot's motion to infer its intent, the subjective benefits of legibility may depend on attention to that motion, even as its effects on coordination persist.

\subsection{Design implications}

From a design perspective, our findings suggest that SRN systems may benefit from prioritizing interaction-level representations of intent over destination inference in constrained hallway-like environments, and from incorporating adaptive mechanisms that allow expressed intent to evolve during interaction. Treating human attention as an explicit contextual variable further underscores the importance of measuring and reasoning about distraction when designing navigation behaviors that balance efficiency with legibility. Importantly, all of these capabilities can be realized within the same control framework, making the balance between functional objectives and interaction-level legibility a directly tunable design parameter.

\subsection{Limitations and future work}

Our studies were conducted in a controlled laboratory hallway with scripted tasks and one-on-one encounters, ensuring precision and repeatability. Ongoing work involves transferring our system to a field deployment in our academic building to test more natural interactions with crowds. Although randomizing the passing side in PL partly separated intent signaling from adherence to the prevailing side convention, legibility and social norms remained inherently intertwined. Future studies could vary legibility formulations and norm compliance independently to examine their individual and combined effects.
For safety and consistency, the robot operated at a fixed maximum speed of $0.33~\text{m/s}$ across all trials, corresponding to the manufacturer’s recommended limit. This choice aligns with prior indoor navigation studies in constrained spaces~\citep{smart2013,Butler2001,Gines2019,Kamezaki2020} and reflects realistic operating speeds of mobile robots deployed in environments such as offices, hospitals, and supermarkets. This likely enabled users to focus more attention on the distraction task without fear of collisions. Because speed was held fixed, our findings characterize legible motion at a single operating point. Future work involves the use of a faster robot, equipped with passive and active safety modules to be safer and more reliable around people.

A broader motivation for our work is to understand when a robot should prioritize legible signals. We approached this question focusing on distraction, suggesting that the benefits of legible motion may persist even when pedestrians are inattentive. Future work will focus on online estimation and adaptation of $\lambda$ to dynamically balance efficiency and legibility. Finally, we will explore additional contextual factors including environmental complexity, physical effort, and task urgency.

\section{Conclusion}

This work examined how robot motion can convey intent during head-on hallway encounters. In Study~1, signaling a passing side reduced Human AA relative to the non-legible baseline, while the adaptive passing-side strategies also received higher competence ratings. The destination-based strategy elicited greater workload and Human AA than the adaptive strategies. In this setting, the robot's near-term passing behavior was more useful for coordination than its final destination alone. Study~2 examined these strategies while pedestrians performed a distraction task. SM continued to reduce Human AA relative to NL, although subjective ratings did not significantly distinguish the algorithms under distraction. Together, the studies support designing social navigation around clear near-term motion and adaptation to pedestrian movement, and evaluating its effects through both human perceptions and observed coordination. Future studies should test other encounter geometries and passing norms while directly measuring how pedestrians infer robot intent.

%% file: references.bib
@article{CSIBRA200760,
  author  = {Gergely Csibra and Gy{\"o}rgy Gergely},
  title   = {`Obsessed with goals': Functions and mechanisms of teleological interpretation of actions in humans},
  journal = {Acta Psychologica},
  volume  = {124},
  number  = {1},
  pages   = {60--78},
  year    = {2007}
}

@inproceedings{Goyal2026LegibilityHRICompanion,
  author    = {Goyal, Pranav and Stratton, Andrew and Mavrogiannis, Christoforos},
  title     = {Rethinking Legibility in Social Robot Navigation: Impact of Intent Representation and Human Attention},
  booktitle = {Companion Proceedings of the ACM/IEEE International Conference on Human-Robot Interaction (HRI)},
  year      = {2026}
}

@book{goffman,
  author    = {Goffman, Erving},
  title     = {Behavior in Public Places: Notes on the Social Organization of Gatherings},
  publisher = {Free Press},
  year      = {1966}
}

@book{goffman2009relations,
  author    = {Goffman, Erving},
  title     = {Relations in Public},
  publisher = {Transaction Publishers},
  year      = {2009}
}

@book{karp1977being,
  author    = {David A. Karp and Gregory Prentice Stone and William C. Yoels},
  title     = {Being urban: a social psychological view of city life},
  publisher = {D. C. Heath},
  year      = {1977}
}

@article{BAKER2009329,
  author  = {Chris L. Baker and Rebecca Saxe and Joshua B. Tenenbaum},
  title   = {Action understanding as inverse planning},
  journal = {Cognition},
  volume  = {113},
  number  = {3},
  pages   = {329--349},
  year    = {2009}
}

@inproceedings{legiblesafety,
  author    = {Lichtenth{\"a}ler, Christina and Lorenz, Tamara and Kirsch, Alexandra},
  title     = {Influence of legibility on perceived safety in a virtual human-robot path crossing task},
  booktitle = {Proceedings of the IEEE International Symposium on Robot and Human Interactive Communication (RO-MAN)},
  year      = {2012}
}

@article{Wolfinger95,
  author  = {Wolfinger, Nicholas H.},
  title   = {Passing Moments: Some Social Dynamics of Pedestrian Interaction},
  journal = {Journal of Contemporary Ethnography},
  volume  = {24},
  number  = {3},
  pages   = {323--340},
  year    = {1995}
}

@inproceedings{mallstudy,
  author    = {Satake, Satoru and Kanda, Takayuki and Glas, Dylan F. and Imai, Michita and Ishiguro, Hiroshi and Hagita, Norihiro},
  title     = {How to approach humans? strategies for social robots to initiate interaction},
  booktitle = {Proceedings of the ACM/IEEE International Conference on Human-Robot Interaction (HRI)},
  year      = {2009}
}

@inproceedings{Dragan,
  author    = {Dragan, Anca D. and Lee, Kenton C. T. and Srinivasa, Siddhartha S.},
  title     = {Legibility and predictability of robot motion},
  booktitle = {Proceedings of the ACM/IEEE International Conference on Human-Robot Interaction (HRI)},
  year      = {2013}
}

@inproceedings{Dragan2013GeneratingLM,
  author    = {Anca D. Dragan and Siddhartha S. Srinivasa},
  title     = {Generating Legible Motion},
  booktitle = {Proceedings of Robotics: Science and Systems (RSS)},
  year      = {2013}
}

@article{trautmanijrr,
  author  = {Peter Trautman and Jeremy Ma and Richard M. Murray and Andreas Krause},
  title   = {Robot navigation in dense human crowds: Statistical models and experimental studies of human-robot cooperation},
  journal = {The International Journal of Robotics Research},
  volume  = {34},
  number  = {3},
  pages   = {335--356},
  year    = {2015}
}

@article{bodden,
  author  = {Christopher Bodden and Daniel Rakita and Bilge Mutlu and Michael Gleicher},
  title   = {A flexible optimization-based method for synthesizing intent-expressive robot arm motion},
  journal = {The International Journal of Robotics Research},
  volume  = {37},
  number  = {11},
  pages   = {1376--1394},
  year    = {2018}
}

@inproceedings{Bastarache,
  author    = {Bastarache, Jean-Luc and Nielsen, Christopher and Smith, Stephen L.},
  title     = {On Legible and Predictable Robot Navigation in Multi-Agent Environments},
  booktitle = {Proceedings of the IEEE International Conference on Robotics and Automation (ICRA)},
  year      = {2023}
}

@article{mavrogiannis2022social,
  author  = {Mavrogiannis, Christoforos and Alves-Oliveira, Patr{\'\i}cia and Thomason, Wil and Knepper, Ross A.},
  title   = {Social momentum: Design and evaluation of a framework for socially competent robot navigation},
  journal = {ACM Transactions on Human-Robot Interaction},
  volume  = {11},
  number  = {2},
  pages   = {1--37},
  year    = {2022}
}

@inproceedings{mavrogiannis2018social,
  author    = {Mavrogiannis, Christoforos I. and Thomason, Wil B. and Knepper, Ross A.},
  title     = {Social Momentum: A Framework for Legible Navigation in Dynamic Multi-Agent Environments},
  booktitle = {Proceedings of the ACM/IEEE International Conference on Human-Robot Interaction (HRI)},
  year      = {2018}
}

@inproceedings{dugas2020ian,
  author    = {Dugas, Daniel and Nieto, Juan and Siegwart, Roland and Chung, Jen Jen},
  title     = {IAN: Multi-behavior navigation planning for robots in real, crowded environments},
  booktitle = {Proceedings of the IEEE/RSJ International Conference on Intelligent Robots and Systems (IROS)},
  year      = {2020}
}

@inproceedings{hart2014gesture,
  author    = {Hart, Justin W. and Gleeson, Brian and Pan, Matthew and Moon, AJung and MacLean, Karon and Croft, Elizabeth},
  title     = {Gesture, gaze, touch, and hesitation: Timing cues for collaborative work},
  booktitle = {Proceedings of the HRI Workshop on Timing in Human-Robot Interaction},
  year      = {2014}
}

@inproceedings{angelopoulos2022familiar,
  author    = {Angelopoulos, Georgios and Vigni, Francesco and Rossi, Alessandra and Russo, Giuseppina and Turco, Mario and Rossi, Silvia},
  title     = {Familiar acoustic cues for legible service robots},
  booktitle = {Proceedings of the IEEE International Conference on Robot and Human Interactive Communication (RO-MAN)},
  year      = {2022}
}

@article{mavrogiannis2023winding,
  author  = {Mavrogiannis, Christoforos and Balasubramanian, Krishna and Poddar, Sriyash and Gandra, Anush and Srinivasa, Siddhartha S.},
  title   = {Winding Through: Crowd Navigation via Topological Invariance},
  journal = {IEEE Robotics and Automation Letters},
  volume  = {8},
  number  = {1},
  pages   = {121--128},
  year    = {2023}
}

@inproceedings{Busch_experiment,
  author    = {Stulp, Freek and Grizou, Jonathan and Busch, Baptiste and Lopes, Manuel},
  title     = {Facilitating intention prediction for humans by optimizing robot motions},
  booktitle = {Proceedings of the IEEE/RSJ International Conference on Intelligent Robots and Systems (IROS)},
  year      = {2015}
}

@inproceedings{RoSAS,
  author    = {Carpinella, Colleen M. and Wyman, Alisa B. and Perez, Michael A. and Stroessner, Steven J.},
  title     = {The Robotic Social Attributes Scale ({RoSAS}): Development and Validation},
  booktitle = {Proceedings of the ACM/IEEE International Conference on Human-Robot Interaction (HRI)},
  year      = {2017}
}

@article{core-challenges2021,
  author  = {Mavrogiannis, Christoforos and Baldini, Francesca and Wang, Allan and Zhao, Dapeng and Trautman, Pete and Steinfeld, Aaron and Oh, Jean},
  title   = {Core Challenges of Social Robot Navigation: A Survey},
  journal = {ACM Transactions on Human-Robot Interaction},
  volume  = {12},
  number  = {3},
  pages   = {1--39},
  year    = {2023}
}

@article{stratton2024characterizingcomplexitysocialrobot,
  author  = {Andrew Stratton and Kris Hauser and Christoforos Mavrogiannis},
  title   = {Characterizing the Complexity of Social Robot Navigation Scenarios},
  journal = {IEEE Robotics and Automation Letters},
  volume  = {10},
  number  = {1},
  pages   = {184--191},
  year    = {2025}
}

@inproceedings{taylor2022observer,
  author    = {Taylor, Ada V. and Mamantov, Ellie and Admoni, Henny},
  title     = {Observer-Aware Legibility for Social Navigation},
  booktitle = {Proceedings of the IEEE International Conference on Robot and Human Interactive Communication (RO-MAN)},
  year      = {2022}
}

@inproceedings{christudy,
  author    = {Mavrogiannis, Christoforos and Hutchinson, Alena M. and Macdonald, John and Alves-Oliveira, Patr{\'\i}cia and Knepper, Ross A.},
  title     = {Effects of Distinct Robot Navigation Strategies on Human Behavior in a Crowded Environment},
  booktitle = {Proceedings of the ACM/IEEE International Conference on Human-Robot Interaction (HRI)},
  year      = {2019}
}

@incollection{hart1988development,
  author    = {Hart, Sandra G. and Staveland, Lowell E.},
  title     = {Development of NASA-TLX (Task Load Index): Results of empirical and theoretical research},
  booktitle = {Human Mental Workload},
  editor    = {Hancock, Peter A. and Meshkati, Najmedin},
  series    = {Advances in Psychology},
  volume    = {52},
  publisher = {North-Holland},
  pages     = {139--183},
  year      = {1988}
}

@inproceedings{williams,
  author    = {Williams, Grady and Wagener, Nolan and Goldfain, Brian and Drews, Paul and Rehg, James M. and Boots, Byron and Theodorou, Evangelos A.},
  title     = {Information theoretic MPC for model-based reinforcement learning},
  booktitle = {Proceedings of the IEEE International Conference on Robotics and Automation (ICRA)},
  year      = {2017}
}

@inproceedings{kirsch,
  author    = {Kruse, Thibault and Basili, Patrizia and Glasauer, Stefan and Kirsch, Alexandra},
  title     = {Legible robot navigation in the proximity of moving humans},
  booktitle = {Proceedings of the IEEE Workshop on Advanced Robotics and its Social Impacts (ARSO)},
  year      = {2012}
}

@inproceedings{Dragan2015study,
  author    = {Dragan, Anca D. and Bauman, Shira and Forlizzi, Jodi and Srinivasa, Siddhartha S.},
  title     = {Effects of Robot Motion on Human-Robot Collaboration},
  booktitle = {Proceedings of the ACM/IEEE International Conference on Human-Robot Interaction (HRI)},
  year      = {2015}
}

@article{Legibilitydiffuser,
  author  = {Bronars, Matthew and Cheng, Shuo and Xu, Danfei},
  title   = {Legibility Diffuser: Offline Imitation for Intent Expressive Motion},
  journal = {IEEE Robotics and Automation Letters},
  volume  = {9},
  number  = {11},
  pages   = {10161--10168},
  year    = {2024}
}

@inproceedings{sacadrl,
  author    = {Chen, Yu Fan and Everett, Michael and Liu, Miao and How, Jonathan P.},
  title     = {Socially aware motion planning with deep reinforcement learning},
  booktitle = {Proceedings of the IEEE/RSJ International Conference on Intelligent Robots and Systems (IROS)},
  year      = {2017}
}

@article{principles,
  author  = {Francis, Anthony and P{\'e}rez-D'Arpino, Claudia and Li, Chengshu and Xia, Fei and Alahi, Alexandre and Alami, Rachid and Bera, Aniket and Biswas, Abhijat and Biswas, Joydeep and Chandra, Rohan and Chiang, Hao-Tien Lewis and Everett, Michael and Ha, Sehoon and Hart, Justin and How, Jonathan P. and Karnan, Haresh and Lee, Tsang-Wei Edward and Manso, Luis J. and Mirsky, Reuth and Pirk, S{\"o}ren and Singamaneni, Phani Teja and Stone, Peter and Taylor, Ada V. and Trautman, Peter and Tsoi, Nathan and V{\'a}zquez, Marynel and Xiao, Xuesu and Xu, Peng and Yokoyama, Naoki and Toshev, Alexander and Mart{\'\i}n-Mart{\'\i}n, Roberto},
  title   = {Principles and Guidelines for Evaluating Social Robot Navigation Algorithms},
  journal = {ACM Transactions on Human-Robot Interaction},
  volume  = {14},
  number  = {2},
  pages   = {1--65},
  year    = {2025}
}

@article{Che_2020,
  author  = {Che, Yuhang and Okamura, Allison M. and Sadigh, Dorsa},
  title   = {Efficient and Trustworthy Social Navigation via Explicit and Implicit Robot--Human Communication},
  journal = {IEEE Transactions on Robotics},
  volume  = {36},
  number  = {3},
  pages   = {692--707},
  year    = {2020}
}

@article{TEB,
  author  = {R{\"o}smann, Christoph and Hoffmann, Frank and Bertram, Torsten},
  title   = {Integrated online trajectory planning and optimization in distinctive topologies},
  journal = {Robotics and Autonomous Systems},
  volume  = {88},
  pages   = {142--153},
  year    = {2017}
}

@inproceedings{ORCA,
  author    = {van den Berg, Jur and Guy, Stephen J. and Lin, Ming and Manocha, Dinesh},
  title     = {Reciprocal n-Body Collision Avoidance},
  booktitle = {Proceedings of the International Symposium of Robotics Research (ISRR)},
  year      = {2011}
}

@article{kim_online_study,
  author  = {Kim, Lawrence H. and Follmer, Sean},
  title   = {Generating Legible and Glanceable Swarm Robot Motion through Trajectory, Collective Behavior, and Pre-attentive Processing Features},
  journal = {ACM Transactions on Human-Robot Interaction},
  volume  = {10},
  number  = {3},
  pages   = {1--25},
  year    = {2021}
}

@inproceedings{szafair_online_study,
  author    = {Szafir, Daniel and Mutlu, Bilge and Fong, Terrence},
  title     = {Communication of intent in assistive free flyers},
  booktitle = {Proceedings of the ACM/IEEE International Conference on Human-Robot Interaction (HRI)},
  year      = {2014}
}

@article{gershman2015computational,
  author  = {Samuel J. Gershman and Eric J. Horvitz and Joshua B. Tenenbaum},
  title   = {Computational rationality: A converging paradigm for intelligence in brains, minds, and machines},
  journal = {Science},
  volume  = {349},
  number  = {6245},
  pages   = {273--278},
  year    = {2015}
}

@article{griffiths2015rational,
  author  = {Griffiths, Thomas L. and Lieder, Falk and Goodman, Noah D.},
  title   = {Rational Use of Cognitive Resources: Levels of Analysis Between the Computational and the Algorithmic},
  journal = {Topics in Cognitive Science},
  volume  = {7},
  number  = {2},
  pages   = {217--229},
  year    = {2015}
}

@article{Wickens2008,
  author  = {Christopher D. Wickens},
  title   = {Multiple Resources and Mental Workload},
  journal = {Human Factors},
  volume  = {50},
  number  = {3},
  pages   = {449--455},
  year    = {2008}
}

@article{strayercell2001,
  author  = {Strayer, David L. and Drews, Frank A.},
  title   = {Cell-Phone--Induced Driver Distraction},
  journal = {Current Directions in Psychological Science},
  volume  = {16},
  number  = {3},
  pages   = {128--131},
  year    = {2007}
}

@article{Horrey2006,
  author  = {Horrey, William J. and Wickens, Christopher D. and Consalus, Kyle P.},
  title   = {Modeling drivers' visual attention allocation while interacting with in-vehicle technologies},
  journal = {Journal of Experimental Psychology: Applied},
  volume  = {12},
  number  = {2},
  pages   = {67--78},
  year    = {2006}
}

@inproceedings{temp_bounded_rationality,
  author    = {Wang, Yiwei and Shintre, Pallavi and Amatya, Sunny and Zhang, Wenlong},
  title     = {Bounded Rational Game-theoretical Modeling of Human Joint Actions with Incomplete Information},
  booktitle = {Proceedings of the IEEE/RSJ International Conference on Intelligent Robots and Systems (IROS)},
  year      = {2022}
}

@inproceedings{Xu2024,
  author    = {Xu, Junhong and Pushp, Durgakant and Yin, Kai and Liu, Lantao},
  title     = {Decision-Making Among Bounded Rational Agents},
  booktitle = {Proceedings of the International Symposium on Distributed Autonomous Robotic Systems (DARS)},
  year      = {2024}
}

@inproceedings{walker2021corl,
  author    = {Walker, Nick and Mavrogiannis, Christoforos and Srinivasa, Siddhartha S. and Cakmak, Maya},
  title     = {Influencing Behavioral Attributions to Robot Motion During Task Execution},
  booktitle = {Proceedings of the Conference on Robot Learning (CoRL)},
  year      = {2021}
}

@misc{Ortega2015,
  author        = {Pedro A. Ortega and Daniel A. Braun and Justin Dyer and Kee-Eung Kim and Naftali Tishby},
  title         = {Information-Theoretic Bounded Rationality},
  year          = {2015},
  eprint        = {1512.06789},
  archiveprefix = {arXiv},
  primaryclass  = {stat.ML}
}

@inproceedings{SunM-RSS-21,
  author    = {Muchen Sun and Francesca Baldini and Peter Trautman and Todd Murphey},
  title     = {Move Beyond Trajectories: Distribution Space Coupling for Crowd Navigation},
  booktitle = {Proceedings of Robotics: Science and Systems (RSS)},
  year      = {2021}
}

@article{Braun2015,
  author    = {Genewein, Tim and Leibfried, Felix and Grau-Moya, Jordi and Braun, Daniel A.},
  title     = {Bounded Rationality, Abstraction, and Hierarchical Decision-Making: An Information-Theoretic Optimality Principle},
  journal   = {Frontiers in Robotics and AI},
  volume    = {2},
  articleno = {27},
  year      = {2015}
}

@inproceedings{Fujioka2024,
  author    = {Fujioka, Yusuke and Liu, Yuyi and Kanda, Takayuki},
  title     = {I Need to Pass Through! Understandable Robot Behavior for Passing Interaction in Narrow Environment},
  booktitle = {Proceedings of the ACM/IEEE International Conference on Human-Robot Interaction (HRI)},
  year      = {2024}
}

@article{pedestrian_distraction_gait_study,
  author    = {Bazzi, Hassan and Cacace, Anthony T.},
  title     = {Altered gait parameters in distracted walking: a bio-evolutionary and prognostic health perspective on passive listening and active responding during cell phone use},
  journal   = {Frontiers in Integrative Neuroscience},
  volume    = {17},
  articleno = {1135495},
  year      = {2023}
}

@article{DistractionFromSmartphonesChangedPedestriansWalkingBehaviorsinOpenAreas,
  author  = {Yue Luo and Nicolas Grimaldi and Haolan Zheng and Wayne C. W. Giang and Boyi Hu},
  title   = {Distraction From Smartphones Changed Pedestrians' Walking Behaviors in Open Areas},
  journal = {Motor Control},
  volume  = {27},
  number  = {2},
  pages   = {275--292},
  year    = {2023}
}

@inproceedings{stefanos_visiblity,
  author    = {Nikolaidis, Stefanos and Dragan, Anca and Srinivasa, Siddhartha},
  title     = {Viewpoint-based legibility optimization},
  booktitle = {Proceedings of the ACM/IEEE International Conference on Human-Robot Interaction (HRI)},
  year      = {2016}
}

@misc{wallkotter2022newapproachevaluatinglegibility,
  author        = {Sebastian Wallk{\"o}tter and Mohamed Chetouani and Ginevra Castellano},
  title         = {A new approach to evaluating legibility: Comparing legibility frameworks using framework-independent robot motion trajectories},
  year          = {2022},
  eprint        = {2201.05765},
  archiveprefix = {arXiv},
  primaryclass  = {cs.RO}
}

@inproceedings{Pacchierotti2006,
  author    = {Pacchierotti, Elena and Christensen, Henrik I. and Jensfelt, Patric},
  title     = {Evaluation of Passing Distance for Social Robots},
  booktitle = {Proceedings of the IEEE International Symposium on Robot and Human Interactive Communication (RO-MAN)},
  year      = {2006}
}

@inproceedings{Pacchierotti2005,
  author    = {Pacchierotti, Elena and Christensen, Henrik I. and Jensfelt, Patric},
  title     = {Embodied Social Interaction for Service Robots in Hallway Environments},
  booktitle = {Proceedings of the International Conference on Field and Service Robotics (FSR)},
  year      = {2006}
}

@inproceedings{smart2013,
  author    = {Lu, David V. and Smart, William D.},
  title     = {Towards more efficient navigation for robots and humans},
  booktitle = {Proceedings of the IEEE/RSJ International Conference on Intelligent Robots and Systems (IROS)},
  year      = {2013}
}

@inproceedings{Kollmitz2015TimeDP,
  author    = {Marina Kollmitz and Kaijen Hsiao and Johannes Gaa and Wolfram Burgard},
  title     = {Time dependent planning on a layered social cost map for human-aware robot navigation},
  booktitle = {Proceedings of the European Conference on Mobile Robots (ECMR)},
  year      = {2015}
}

@inproceedings{liu2023intentionawarerobotcrowd,
  author    = {Shuijing Liu and Peixin Chang and Zhe Huang and Neeloy Chakraborty and Kaiwen Hong and Weihang Liang and D. Livingston McPherson and Junyi Geng and Katherine Driggs-Campbell},
  title     = {Intention Aware Robot Crowd Navigation with Attention-Based Interaction Graph},
  booktitle = {Proceedings of the IEEE International Conference on Robotics and Automation (ICRA)},
  year      = {2023}
}

@inproceedings{singamaneni2021humanawarenavigationplannerdiverse,
  author    = {Phani Teja Singamaneni and Anthony Favier and Rachid Alami},
  title     = {Human-Aware Navigation Planner for Diverse Human-Robot Interaction Contexts},
  booktitle = {Proceedings of the IEEE/RSJ International Conference on Intelligent Robots and Systems (IROS)},
  year      = {2021}
}

@article{scholler2020constantvelocitymodelteach,
  author  = {Christoph Sch{\"o}ller and Vincent Aravantinos and Florian Lay and Alois Knoll},
  title   = {What the Constant Velocity Model Can Teach Us About Pedestrian Motion Prediction},
  journal = {IEEE Robotics and Automation Letters},
  volume  = {5},
  number  = {2},
  pages   = {1696--1703},
  year    = {2020}
}

@article{Gao2022,
  author    = {Gao, Yuxiang and Huang, Chien-Ming},
  title     = {Evaluation of Socially-Aware Robot Navigation},
  journal   = {Frontiers in Robotics and AI},
  volume    = {8},
  articleno = {721317},
  year      = {2022}
}

@inproceedings{Guzzi2013,
  author    = {Guzzi, J{\'e}r{\^o}me and Giusti, Alessandro and Gambardella, Luca M. and Theraulaz, Guy and Di Caro, Gianni A.},
  title     = {Human-friendly robot navigation in dynamic environments},
  booktitle = {Proceedings of the IEEE International Conference on Robotics and Automation (ICRA)},
  year      = {2013}
}

@inproceedings{knepper_hri_2017,
  author    = {Knepper, Ross A. and Mavrogiannis, Christoforos I. and Proft, Julia and Liang, Claire},
  title     = {Implicit Communication in a Joint Action},
  booktitle = {Proceedings of the ACM/IEEE International Conference on Human-Robot Interaction (HRI)},
  year      = {2017}
}

@article{FARIA2024104107,
  author    = {Miguel Faria and Francisco S. Melo and Ana Paiva},
  title     = {``Guess what I'm doing'': Extending legibility to sequential decision tasks},
  journal   = {Artificial Intelligence},
  volume    = {330},
  articleno = {104107},
  year      = {2024}
}

@inproceedings{cuan2022gesture2pathimitationlearninggestureaware,
  author    = {Catie Cuan and Tsang-Wei Edward Lee and Emre Fisher and Anthony Francis and Leila Takayama and Tingnan Zhang and Alexander Toshev and S{\"o}ren Pirk},
  title     = {Gesture2Path: Imitation Learning for Gesture-Aware Navigation},
  booktitle = {Proceedings of the International Conference on Social Robotics (ICSR)},
  year      = {2025}
}

@article{kathuria2025learningimplicitsocialnavigation,
  author  = {Tribhi Kathuria and Ke Liu and Junwoo Jang and X. Jessie Yang and Maani Ghaffari},
  title   = {Learning Implicit Social Navigation Behavior Using Deep Inverse Reinforcement Learning},
  journal = {IEEE Robotics and Automation Letters},
  volume  = {10},
  number  = {5},
  pages   = {5146--5153},
  year    = {2025}
}

@inproceedings{lichtenthaler:hal-01684307,
  author    = {Lichtenth{\"a}ler, Christina and Kirsch, Alexandra},
  title     = {Towards Legible Robot Navigation - How to Increase the Intend Expressiveness of Robot Navigation Behavior},
  booktitle = {Proceedings of the ICSR Workshop on Embodied Communication of Goals and Intentions},
  year      = {2013}
}

@inproceedings{Amirian2024,
  author    = {Amirian, Javad and Abrini, Mouad and Chetouani, Mohamed},
  title     = {Legibot: Generating Legible Motions for Service Robots Using Cost-Based Local Planners},
  booktitle = {Proceedings of the IEEE International Conference on Robot and Human Interactive Communication (RO-MAN)},
  year      = {2024}
}

@inproceedings{Tsoi2024,
  author    = {Tsoi, Nathan and Romero, Jessica and V{\'a}zquez, Marynel},
  title     = {How Do Robot Experts Measure the Success of Social Robot Navigation?},
  booktitle = {Companion Proceedings of the ACM/IEEE International Conference on Human-Robot Interaction (HRI)},
  year      = {2024}
}

@article{Biswas2022,
  author  = {Biswas, Abhijat and Wang, Allan and Silvera, Gustavo and Steinfeld, Aaron and Admoni, Henny},
  title   = {SocNavBench: A Grounded Simulation Testing Framework for Evaluating Social Navigation},
  journal = {ACM Transactions on Human-Robot Interaction},
  volume  = {11},
  number  = {3},
  pages   = {1--24},
  year    = {2022}
}

@article{Zhang2022,
  author  = {Bingqing Zhang and Javad Amirian and Harry Eberle and Julien Pettr{\'e} and Catherine Holloway and Tom Carlson},
  title   = {From HRI to CRI: Crowd Robot Interaction---Understanding the Effect of Robots on Crowd Motion},
  journal = {International Journal of Social Robotics},
  volume  = {14},
  number  = {3},
  pages   = {631--643},
  year    = {2022}
}

@article{Butler2001,
  author  = {John Travis Butler and Arvin Agah},
  title   = {Psychological Effects of Behavior Patterns of a Mobile Personal Robot},
  journal = {Autonomous Robots},
  volume  = {10},
  number  = {2},
  pages   = {185--202},
  year    = {2001}
}

@article{Gines2019,
  author    = {Gin{\'e}s, Jonatan and Mart{\'\i}n, Francisco and Vargas, David and Rodr{\'\i}guez, Francisco J. and Matell{\'a}n, Vicente},
  title     = {Social Navigation in a Cognitive Architecture Using Dynamic Proxemic Zones},
  journal   = {Sensors},
  volume    = {19},
  number    = {23},
  articleno = {5189},
  year      = {2019}
}

@article{Kamezaki2020,
  author  = {Mitsuhiro Kamezaki and Ayano Kobayashi and Yuta Yokoyama and Hayato Yanagawa and Moondeep Shrestha and Shigeki Sugano},
  title   = {A Preliminary Study of Interactive Navigation Framework with Situation-Adaptive Multimodal Inducement: Pass-By Scenario},
  journal = {International Journal of Social Robotics},
  volume  = {12},
  number  = {2},
  pages   = {567--588},
  year    = {2020}
}

@phdthesis{adathesis,
  author = {Taylor, Ada V.},
  title  = {Goal-Expressive Movement for Social Navigation: Where and When to Behave Legibly},
  school = {Carnegie Mellon University},
  year   = {2024}
}

@inproceedings{poddar2023crowdmotionpredictionrobot,
  author    = {Poddar, Sriyash and Mavrogiannis, Christoforos and Srinivasa, Siddhartha S.},
  title     = {From Crowd Motion Prediction to Robot Navigation in Crowds},
  booktitle = {Proceedings of the IEEE/RSJ International Conference on Intelligent Robots and Systems (IROS)},
  year      = {2023}
}

@article{olivier2012minimal,
  author  = {Olivier, Anne-H{\'e}l{\`e}ne and Marin, Antoine and Cr{\'e}tual, Armel and Pettr{\'e}, Julien},
  title   = {Minimal predicted distance: A common metric for collision avoidance during pairwise interactions between walkers},
  journal = {Gait \& Posture},
  volume  = {36},
  number  = {3},
  pages   = {399--404},
  year    = {2012}
}

@article{vassallo2017how,
  author  = {Vassallo, Christian and Olivier, Anne-H{\'e}l{\`e}ne and Sou{\`e}res, Philippe and Cr{\'e}tual, Armel and Stasse, Olivier and Pettr{\'e}, Julien},
  title   = {How do walkers avoid a mobile robot crossing their way?},
  journal = {Gait \& Posture},
  volume  = {51},
  pages   = {97--103},
  year    = {2017}
}

@inproceedings{probst2025responsibility,
  author    = {Probst, Malte and Wenzel, Raphael and Dasi, Monica},
  title     = {Responsibility and Engagement -- Evaluating Interactions in Social Robot Navigation},
  booktitle = {Proceedings of the IEEE International Conference on Robotics and Automation (ICRA)},
  year      = {2026}
}

@article{neggers2022speed,
  author    = {Neggers, Margot M. E. and Cuijpers, Raymond H. and Ruijten, Peter A. M. and IJsselsteijn, Wijnand A.},
  title     = {The effect of robot speed on comfortable passing distances},
  journal   = {Frontiers in Robotics and AI},
  volume    = {9},
  articleno = {915972},
  year      = {2022}
}

@inproceedings{cathcart2023opinion,
  author    = {Cathcart, Charlotte and Santos, Mar{\'\i}a and Park, Shinkyu and Leonard, Naomi Ehrich},
  title     = {Proactive Opinion-Driven Robot Navigation Around Human Movers},
  booktitle = {Proceedings of the IEEE/RSJ International Conference on Intelligent Robots and Systems (IROS)},
  year      = {2023}
}

@inproceedings{geldenbott2024legible,
  author    = {Geldenbott, Jasper and Leung, Karen},
  title     = {Legible and Proactive Robot Planning for Prosocial Human--Robot Interactions},
  booktitle = {Proceedings of the IEEE/RSJ International Conference on Intelligent Robots and Systems (IROS)},
  year      = {2024}
}
